\documentclass[11pt]{article}
\pdfoutput=1

\def\red#1{}

\usepackage[dvipsnames]{xcolor}

\usepackage{diagbox}
\def\beq{\begin{equation} }\def\eeq{\end{equation} }\def\1{\mathbf{1}}

\usepackage[framemethod=default]{mdframed}
\usepackage{caption}
\usepackage{indentfirst}
\usepackage{bm, mathrsfs, graphics,float,amssymb,amsmath,subeqnarray,setspace,graphicx,amsthm,epstopdf,subfigure, enumerate, color, array, booktabs}
\usepackage[utf8]{inputenc}
\usepackage{natbib}

\usepackage{fullpage}
\numberwithin{equation}{section}

\newtheorem{lemma}{Lemma}
\newtheorem{theorem}{Theorem}
\newtheorem{proposition}{Proposition}
\newtheorem{definition}{Definition}
\newtheorem{corollary}[theorem]{Corollary}
\newtheorem{remark}{Remark}

\ifx\assumption\undefined
\newtheorem{assumption}{Assumption}
\fi

\newcommand{\bSigma}{\bm{\Sigma}}

\newcommand{\cO}{\mathcal{O}}

\newcommand{\RR}{\mathbb{R}}

\newcommand{\I}{\mathbf{I}}

\newcommand{\cN}{{\mathcal{N}}}

\newcommand{\PP}{\mathbb{P}}

\newcommand{\tR}{{\tilde{\R}}}
\newcommand{\Sc}{{\S_C}}
\newcommand{\Sr}{\S_R}
\newcommand{\Si}{{\bf \Sigma}}
\newcommand{\tC}{\tilde{\C}}
\def\Ps{\mbox{\boldmath$\Psi$\unboldmath}}
\def\Ome{\mbox{\boldmath$\Omega$\unboldmath}}

\def\TX{\tilde{\X}}
\def\HX{\hat{\X}}
\def\rk{\mathrm{rank}}
\def\sym{\mathrm{sym}}
\def\Pii{\mbox{\boldmath$\Pi$\unboldmath}}
\def\HB{\mathbb{H}}
\def\SF{{\bf{SF}}}
\newcommand{\argmin}{\mathop{\mathrm{argmin}}}

\usepackage{bbm}

\def\A{{\bf A}}

\def\B{{\bf B}}

\def\C{{\bf C}}

\def\D{{\bf D}}

\def\G{{\bf G}}
\def\H{{\bf H}}
\def\I{{\bf I}}
\def\K{{\bf K}}

\def\M{{\bf M}}

\def\N{{\bf N}}

\def\PP{{\bf P}}

\def\R{{\bf R}}
\def\S{{\bf S}}
\def\s{{\bf s}}
\def\T{{\bf T}}

\def\U{{\bf U}}
\def\u{{\bf u}}
\def\V{{\bf V}}

\def\X{{\bf X}}
\def\x{{\bf x}}

\def\Z{{\bf Z}}

\newcommand{\tA}{{\tilde{\A}}}
\newcommand{\nnz}{{\mathrm{nnz}}}

\def\TH{\tilde{H}}
\newcommand{\ti}[1]{\tilde{#1}}
\def\diag{\mathrm{diag}}
\newcommand{\norm}[1]{\left\|#1\right\|}

\def\tr{\mathrm{tr}}
\def\TimeSketch{{T_{\mathrm{sketch}}}}

\def\RBmn{\RR^{m\times n}}

\def\nystrom{{Nystr\"{o}m }}

\usepackage{multirow}
\usepackage{tablefootnote}
\usepackage{colortbl}
\usepackage{hhline}

\usepackage{algorithm}
\usepackage{algorithmic}

\begin{document}
\title{
 Fast Generalized Matrix Regression with Applications in Machine Learning
}

\author{
Haishan Ye
\thanks{
Shenzhen Research Institute of Big Data;
email: hsye\_cs@outlook.com;
}
\and
Shusen Wang
\thanks{
	Stevens Institute of Technology;
	email:Shusen.Wang@stevens.edu
}
\and
Zhihua Zhang
\thanks{
Peking University;
email: zhzhang@math.pku.edu.cn
}
\and
Tong Zhang
\thanks{
Hong Kong University of Science and Technology; 
email: tongzhang@tongzhang-ml.org
}
}
\date{\today}

\maketitle
\begin{abstract}
	Fast matrix algorithms have become the fundamental tools of machine learning in big data era.
	The generalized matrix regression problem is widely used in the matrix approximation such as CUR decomposition, kernel matrix approximation, and stream singular value decomposition (SVD), etc.
	In this paper, we  propose a fast generalized matrix regression algorithm (Fast GMR) which utilizes sketching technique to solve the GMR problem efficiently.
	Given error parameter $0<\epsilon<1$, the Fast GMR algorithm can achieve a $(1+\epsilon)$ relative error with the sketching sizes being of order $\cO(\epsilon^{-1/2})$ for a large group of GMR problems.
	We apply the Fast GMR algorithm to the symmetric positive definite matrix approximation and single pass singular value decomposition and they achieve a better performance than conventional algorithms.
	Our empirical study also validates the effectiveness and efficiency of our proposed algorithms.
\end{abstract}

\section{Introduction}
Matrix manipulations are the basis of modern data analysis. As the datasets becomes larger and larger, it is much more difficult to perform exact matrix multiplication, inversion, and  decomposition. Consequently, matrix approximation techniques have been extensively studied, including approximate matrix multiplication \citep{cohen1999approximating,drineas2006fast,kyrillidis2014approximate}, the least square regression \citep{clarkson2013low,sarlos2006improved} and low-rank matrix approximation \citep{bourgain2013toward,boutsidis2014near,clarkson2013low,martinsson2011randomized}. 
In this paper, we focus on the generalized matrix regression (GMR) problem which is defined as follows
\begin{align}
\X^{\star} = \argmin_{\X}\|\A - \C\X\R\|_F, \label{eq:glsr}
\end{align}
where $\A\in\RR^{m\times n}$, $\C\in\RR^{m\times c}$ and $\R\in \RR^{r\times n}$ are any fixed matrices with $c \ll n$ and $r\ll m$. 
Notation $\norm{\A}_F = \sqrt{\sum_{i,j}{\A_{i,j}^2}}$ is the Frobenius norm. 
Note that $\C$ and $\R$ are any fixed matrices and can be (in)dependent on $\A$.
It is well know that the optimal solution of the problem is $\X^\star = \C^\dagger \A \R^\dagger$, where $\C^\dagger$ is the pseudo inverse of $\C$ and $\R^\dagger$ is similarly defined. 

The GMR problem arises from the randomized low rank approximation. 
For example, in the CUR decomposition, one first tries to find columns $\C$ and and rows $\R$ which maintain the principal information of column and row spaces of $\A$, respectively.
Then, a core matrix $\X^\star = \C^\dagger \A\R\dagger$ (the solution of \eqref{eq:glsr}) is constructed to make the approximation error $\norm{\A - \C\X\R}_F$ as small as possible.  Accordingly,  the CUR decomposition $\A \approx \C\X^\star\R$ is obtained \citep{drineas2008relative,wang15}.
Similarly, in the randomized SVD algorithms, approximate top-$k$ left and right singular vectors $\U$ and $\V$ are obtained by random sketching (projection) method. Then, the core matrix $\X$ are also computed by GMR problem with $\C$ and $\R$ are replaced by $\U$ and $\V$ in \eqref{eq:glsr}, respectively \citep{Clarkson2009Numerical,clarkson2017low}. 

However, it costs $\cO(\nnz(\A)\cdot\min(c,r)+mc^2+nr^2)$ time to solve the GMR exactly to obtain $\X^{*}$, where $\nnz(\A)$ denotes the number of non-zero entries of $\A$.
To solve GMR problem efficiently, we resort to the sketching technique. 
Using two sketching matrices , instead of solving problem~\eqref{eq:glsr}, we are going to solve the following sketched problem
\begin{equation*}
\min_{\X} \norm{\Sc\left(\C\X\R - \A\right)\Sr^T}
\end{equation*}
where $\Sc \in\RR^{s_c\times m}$ and $\Sr \in\RR^{s_r\times n}$ are two sketching matrices with $s_c \ll m$ and $s_r\ll n$.
Then the sketched problem can be solved at the costs of $\cO(s_cs_r\cdot\min(s_c,s_r)+s_cc^2+s_rr^2)$ computation.
We can observe that the computational cost of solving the sketched GMR problem is independent of the input sizes of $\A$.
Furthermore, we will prove that the solution of the sketched GMR problem can well approximate the exact solution of GMR.
Therefore, GMR can be solved efficiently by utilizing the sketching technique.

Another important motivation behind the fast GMR lies in the stream (single pass) low rank matrix approximation \citep{tropp2017practical}.
In this setting, the principal column and row information can be captured by sketching matrices $\tilde{\Ome}\in\RR^{n\times c}$ and $\tilde{\Ps}\in\RR^{r\times m}$ as follows 
\begin{equation*}
\C = \A\tilde{\Ome}, \quad \mbox{and} \quad \R = \tilde{\Ps} \A.
\end{equation*} 
Similar to CUR decomposition, a core matrix is needed to make the low rank approximation error as small as possible.
Recall that the optimal core matrix is $\C^\dagger\A\R^\dagger$. 
However, in the stream setting, the data will be dropped after obtaining $\C$ and $\R$, one can not obtain the entries of $\A$ any more.
To solve the dilemma,  we can resort to the sketching technique again and construct a small matrix $\tA = \Sc\A\Sr^T$ with low space cost.
Then we will use the sketched matrix to compute an approximate core matrix.
The above procedure can be summarized as to solve the sketched GMR problem.
Therefore, the fast GMR algorithm is also an important pillar in stream low rank approximation.

In this paper, we aims to solve the generalized matrix regression problem efficiently by utilizing the sketching technique. 
We summarize our contribution as follows.
\begin{enumerate}
	\item We propose fast generalized matrix regression algorithm which can solve the GMR problem with a $(1+\epsilon)$-relative error bound.
	\item We show that the sketching sizes used in fast generalized matrix regression algorithm is only of order $\epsilon^{-1/2}$ in most cases to achieve a $(1+\epsilon)$-relative error bound. 
	Moreover, this kind of bound of sketching sizes is unknown before.
	\item We apply our fast GMR algorithm to the symmetric positive semi-definite matrix approximation (Algorithm~\ref{alg:psd}). 
	Our method achieves substantially better theoretical and empirical performance than several conventional algorithms.
	\item We apply our Fast GMR algorithm to the single pass SVD decomposition (Algorithm~\ref{alg:Pass_effi_SVD}).
	Our algorithm can obtain an approximate SVD decomposition of the matrix $\A$ with $(1+\epsilon)$-relative error with respect to $\|\A - \A_k\|_F$ with $\A_k$ be the best rank $k$ approximation of $\A$ at the cost of only $\cO(\nnz(\A))$ computation and $\cO((m+n)k/\epsilon)$ storage space.
\end{enumerate}

\paragraph{More Related Literature}

The least square regression (LSR) problem $\min_\X\|\C\X - \A\|_F$ is closely related to the generalized matrix regression. 
Using sketching technique to solve the LSR problem has been widely studied \citep{Clarkson2009Numerical,sarlos2006improved,clarkson2013low}.
Similar to our fast GMR algorithm, one tries to solve the sketched SLR $\min_{\X}\|\S\C\X - \S\A\|_F$ where $\S$ is a sketching matrix of proper sizes.
Compared with the GMR problem, we can observe that GMR has an extra term $\R$ than the LSR problem.
Therefore, the theories of analyzing the sketched LSR can not directly derive the bounds of sketching sizes of the GMR problem.
Furthermore, our theoretical analysis shows that the sketching sizes is only of order $\epsilon^{-1/2}$ is sufficient for a large group of GMR problems.
In contrast, the sketching size of the LSR problem should be at least of order  $\epsilon^{-1}$ which has achieved the lower bound provided in \citep{Clarkson2009Numerical}.

There are several symmetric positive semi-definite (SPSD) matrix approximation have been proposed in which \nystrom is the most popular one \citep{williams2001using}. \nystrom has been widely used in kernel approximation.
Recently,  the fast symmetric positive semi-definite (SPSD) matrix approximation was proposed by \citep{wang15} which applies the sketching technique to reduce the computational cost.
However, this method is not sufficiently efficient. 
Specifically, to achieve a $(1+\epsilon)$-relative error for the kernel matrix approximation given the chosen column matrix $\C\in\RR^{n\times c}$, the sketching size of the method in \citep{wang15} is at least $\cO(cn^{1/2}\epsilon^{-1/2})$ which means that this method needs to compute another $\cO(nc^2\epsilon^{-1})$ entries of the kernel matrix.
In contrast, our method (Algorithm~\ref{alg:psd}) achieves the SPSD approximation in a different way compared with the fast SPSD of \citep{wang15} and only requires to compute $\cO(c^2\epsilon^{-1})$ extra entries in most applications.
Therefore, our algorithm is better than the fast SPSD method.

Recently, \citet{tropp2017practical} brought up a single pass SVD algorithm (Algorithm~\ref{alg:sp_svd_hk} in Appendix) using sketching matrices. For any input matrix $\A\in\RBmn$, target rank $k$ and error parameter $\epsilon$, the sketch sizes of their algorithm are $\cO\left(k/\epsilon\right)$ and $\cO\left(k/\epsilon^2\right)$ to achieve a $(1+\epsilon)$-relative error bound with respect to $\|\A - \A_k\|_F$. 
The algorithm runs in $\cO\left(k/\epsilon\cdot\nnz(\A) + mk^2/\epsilon^2 + n k^2/\epsilon^3+ k^3/\epsilon^4\right)$ time using Gaussian projection. Implementing with a proper sketching matrix, this algorithm can run in input sparsity. 
In fact, the algorithm of \citet{tropp2017practical} is equivalent to the one of \citet{Clarkson2009Numerical} in algebraic essence. Single pass SVD algorithms via sketching  have also been widely studied in other works \citep{clarkson2013low,nelson2013osnap, Cohen2015}.
In contrast to the previous work mentioned above, our single pass algorithm only takes $\cO((m+n)k\epsilon^{-1})$ storage space and $\cO(\nnz(\A))$ computational cost. 

\section{Preliminary}\label{sec:notation}

Section~\ref{subsec:notation} gives the notation used in this paper. Section~\ref{subsec:cost} gives the cost of several basic matrix operations. Section~\ref{subsec:sketch} introduces matrix sketching techniques and several important kinds of sketching matrices. 

\subsection{Notation} \label{subsec:notation}

Let $\I_m$ be the $m{\times}m$ identity matrix. 
Given a matrix $\A=[a_{ij}] \in \RR^{m \times n}$ of rank $ \rho $ and a positive integer $k\leq \rho$, its SVD is given as
$\A=\U\Si\V^{T}=\U_{k} \Si_{k} \V_{k}^{T}+\U_{\rho\setminus k} \Si_{\rho{\setminus} k}\V_{\rho{\setminus}k}^{T}$,
where $\U_{k}$ and $\U_{\rho{\setminus}k}$ contain the left singular vectors of $\A$,  $\V_{k}$ and $\V_{\rho{\setminus}k}$ contain the right
singular vectors of $\A$, and $\Si=\diag(\sigma_1, \ldots, \sigma_{\rho})$ with $\sigma_1\geq \sigma_2 \geq \cdots \geq \sigma_{\rho}>0$ are
the nonzero singular values of $\A$. Accordingly,   $\|\A\|_{F} \triangleq (\sum_{i,j}a_{ij}^{2})^{1/2}=(\sum_{i}\sigma_{i}^{2})^{1/2}$
is the Frobenius norm of $\A$ and
$\|\A\|_{2}\triangleq \sigma_{1}$ is the spectral norm.

Additionally, $\A^{\dagger} \triangleq \V\Si^{-1}\U^{T} \in \RR^{n \times m}$ is
the  Moore-Penrose pseudoinverse of $\A$, which is  unique. It is easy to verify that ${\rk}(\A^{\dagger}) ={\rk}(\A)=\rho$. 
Moreover, for all $i=1, \dots, \rho$,
$\sigma_{i}(\A^{\dagger})=1/\sigma_{\rho-i+1}(\A)$. If  $\A$ is of full row rank, then $\A\A^{\dagger}=\I_{m}$.
Also, if $\A$ is of full column rank, then $\A^{\dagger}\A=\I_{n}$. 
When $m=n$, $\text{tr}(\A) = \sum_{i=1}^{n}a_{ii}$ is the trace of $\A$.

It is well known that $\A_{k}\triangleq \U_{k}\Si_{k}\V_{k}^{T}$ is the minimizer of both $\min \|\A-\X\|_{F}$ and $\min \|\A-\X\|_{2}$ over all matrices $\X \in \RR^{m \times n}$ of rank at most $k\leq\rho$. Thus, $\A_k$ is called the best rank-$k$ approximation of $\A$.

Assume that $\rho \leq n$. The column leverage scores of $\A$ are $\ell_i = \|\V_{i,:}\|_2^2$ for $i=1,\dots,n$. 
It holds that $\sum_{i=1}^{n} \ell_i= \rho$. 
Furthermore, we can define the column coherence $\mu(\A) = \frac{m}{\rho}\max_i\norm{\u_i}^2$.
If $\rho<m$, the row leverage scores and row coherence can be similarly defined.


\subsection{Computation cost of matrix operations} \label{subsec:cost}

We will give the computation cost of basic matrix operations. For matrix multiplication, given dense matrices $\B\in\RR^{m\times n}$ and $\C\in\RR^{n\times k}$, the basic cost of the matrix product $\B\times \C$ is $O(mnk)$ flops. It costs $O(k\cdot\nnz(\B))$ flops for the matrix product $\B\times \C$ when $\B$ is sparse, where $\nnz(\B)$ denotes the number of nonzero entries of $\B$. If $\S$ is a count sketch or OSNAP sketching matrix described in the next subsection \citep{woodruff2014sketching}, then the product $\S\times \B$ costs $O(\nnz(\B))$. 

\subsection{Matrix Sketching}\label{subsec:sketch}
Matrix sketching methods are useful tools for speeding up numerical linear algebra and machine learning 
\citep{woodruff2014sketching,mahoney2011ramdomized}. Popular matrix sketching methods include random sampling \citep{drineas2006sampling,drineas2012fast},  Gaussian projection \citep{halko2011finding,johnson1984extensions}, 
subsampled randomized Hadamard transform (SRHT) \citep{drineas2012fast,halko2011finding}, count sketch \citep{clarkson2013low}, OSNAP \citep{nelson2013osnap}, etc.

\paragraph{Leverage score sampling.}
Define probabilities $p_1 , \cdots , p_m \in (0, 1)$ ($\sum_i p_i = 1$).
A subset of rows sampled from $\A$ according to the probabilities forms a sketch of $\A$.
For $i= 1$ to $m$, the $i$-th row is sampled with probability $p_i$;
once selected, it gets scaled by $\frac{1}{\sqrt{sp_i}}$.
The procedure of row sampling can be equivalently captured by a sketching matrix $\S \in \RR^{s\times m}$.
There is exactly one non-zero entry in each row of $\S$, 
whose position indicates the index of the selected row.
Uniform sampling sets $p_i = \frac{1}{n}$;
leverage score sampling uses $p_i = \tfrac{\ell_i}{\sum_i \ell_i}$, for $i\in[m]$,
where $\ell_i$ is the $i$-th leverage score of some matrix.

\paragraph{Gaussian projection.}
The most classical sketching matrix is the Gaussian random matrix $\G\in\RR^{s\times m}$ 
whose entries are i.i.d.\ sampled from $\cN (0, 1/s)$.
Because of well-known concentration properties of Gaussian random matrices \citep{woodruff2014sketching}, they are very attractive. However, Gaussian random matrices are dense, so it takes $\cO(mns)$ to compute $\G\A$, 
which makes Gaussian projection inefficient.
\paragraph{Subsampled randomized Hadamard transform (SRHT).}
SRHT is usually a more efficient alternative of Gaussian projection for dense matrix. Let $\H_m\in\RR^{m\times m}$ be the Walsh-Hadamard matrix, $\D\in\RR^{m\times m}$ be a diagonal matrix with diagonal entries sampled from $\{+1,-1\}$ uniformly, and $\PP\in\RR^{s\times m}$ be uniform sampling matrix. Then the matrix $\S = \frac{1}{\sqrt{s}}\PP\H\D$ is an SRHT matrix, 
and it costs $\cO(mn\log s)$ time when applied to an $m\times n$	matrix.
\begin{table}[]\setlength{\tabcolsep}{0.5pt}
	\caption{For property 2, the leverage score sampling is w.r.t.\ the row leverage scores of $\A$ and $\A$ is an orthonormal matrix.
	}
	\label{tb:property}
	\begin{center}
		\begin{tabular}{c c c }
			\hline
			{\bf Sketching}             &~~~~Property 1~~~~&~~~~Property 2~~~~\\ \hline
			Leverage Sampling  &~~~$ \frac{k}{\eta^2} \log \frac{k}{\delta_1} $~~~
			&~~~$ \frac{1}{\epsilon^2 \delta_2}  $~~~
			\\
			Gaussian Projection\ &~~~$\frac{  k +  \log (1/\delta_1) }{\eta^2}$~~~
			& ~~~$\frac{1}{\epsilon^2 \delta_2}$~~~\\
			SRHT            &~~~$\frac{k + \log n}{\eta^2} \log \frac{k}{\delta_1}$~~~
			& ~~~$\frac{ \log md}{\epsilon^2  \delta_2  }$~~~\\
			Count Sketch &~~~$\frac{k^2 }{\delta_1 \eta^2}$~~~
			& ~~~$\frac{ 1}{\epsilon^2 \delta_2}$~~~\\
			OSNAP &~~~$\left(\frac{k}{ \eta}\right)^{1+\gamma}\log\frac{1}{\delta_1}$~~~
			& ~~~$\frac{ 1}{\epsilon^2 \delta_2}$~~~\\
			\hline
		\end{tabular}
	\end{center}
\end{table}
\paragraph{Count sketch.}	
The count sketch matrix $\S\in\RR^{s\times m}$ 
has only one non-zero entry in each column;
the entry is a random sign, and its position is uniformly at random \citep{clarkson2013low}. 
Therefore, it is efficient to obtain $\S\A$ especially when $\A$ is sparse.

\paragraph{OSNAP.}
An OSNAP matrix $\S\in\RR^{s\times m}$ is of the form that there is only $p$ non-zero entries uniformly sampling from $\{1,-1\}$ in each column and these non-zero entries are distributed uniformly \citep{nelson2013osnap}.

The sketching matrices have two important properties which we describe in the following lemma.
The first property is known as subspace embedding \citep{woodruff2014sketching}. 
It shows that all singular values of $\S\A$ are close to the ones of $\A$. 
The second property shows that the sketching matrix can preserve the multiplication of two matrices. 
Furthermore, for these four kinds of sketching matrices, the combination of any two kinds of sketching matrices still have the above two properties.

\begin{lemma}\label{lem:ske_prop}
	Let $\A\in\RR^{m\times d}$ be any fixed matrix of rank $k$, $\B\in\RR^{m\times n}$ be another fixed matrix, and $\S\in\RR^{s\times m}$ be any sketching matrix described in this subsection. The order of $s$ is listed in Table~\ref{tb:property}. Then we have  
	\begin{align*}
	(1-\eta)\|\A\x\|_2^2\leq\|\S\A\x\|_2^2 \leq (1+\eta)\|\A\x\|_2^2 \; (\textrm{property 1})
	\end{align*}
	with probability at least $1-\delta_1$,\\
	and 
	\begin{align*}
	\|\B^T\S^T\S\A - \B^T\A\|_F^2 \leq \epsilon^2\|\A\|_F^2\|\B\|_F^2 \quad\text{ {} }(\textrm{property 2})
	\end{align*}
	with probability at least $1-\delta_2$. Here $\eta$ and $\epsilon$ are the error parameters.
\end{lemma}

The properties of leverage score sampling were first proved by  \citet{drineas2006sampling,drineas2008relative,woodruff2014sketching}.  These theoretical results of Gaussian projection can be found in \citep{woodruff2014sketching}. The properties of SRHT were established by \citet{tropp2011improved, drineas2011faster}. Theoretical guarantees of count sketch were proved by \citet{clarkson2013low,meng2013low}.  Theoretical guarantees of OSNAP were then given by \citet{nelson2013osnap}.

\section{Fast Generalized Matrix Regression} \label{sec:glsr}

The fast GMR method is going to utilize the sketching technique to solve the GMR problem efficiently. 
Instead of solving problem~\eqref{eq:glsr} directly, the fast GMR method will solve the following sketched problem 
\begin{equation}
\min_{\X} \bigg\|{\underbrace{\Sc \C}_{\tC}\X \underbrace{\R\Sr^T}_{\tR} - \underbrace{\Sc \A\Sr^T}_{\tA} }\bigg\|. \label{eq:ske_gmlr}
\end{equation}
$\Sc \in\RR^{s_c\times m}$ and $\Sr \in\RR^{s_r\times n}$ are two sketching matrices with $s_c \ll m$ and $s_r\ll n$.
Thus, we will only to solve a GMR problem 
\[
\min_{\X}\norm{\tC\X\tR - \tA}
\]
with much reduced sizes   
which requires much less computational cost compared with the original GMR problem.
Furthermore, if $\tA = \Sc \A\Sr^T$ can be obtained efficiently, then fast GMR method can much reduce the computational cost to solve the GMR problem.
The detailed description of the Fast GMR method is listed in Algorithm~\ref{alg:gmr}.

\begin{algorithm}[tb]
	\caption{The Fast GMR method.}
	\label{alg:gmr}
		\begin{algorithmic}[1]
			\STATE {\bf Input:} Real matrices $\A \in \RR^{m\times n}$, $\C\in\RR^{m\times c}$ and $\R\in\RR^{r\times n}$.  Sketching sizes $s_c$ and $s_r$;
			\STATE Construct the sketching matrices $\Sc\in\RR^{s_c\times m}$ and $\Sr\in\RR^{s_r\times n}$;
			\STATE Compute the sketched matrices $\Sc \C\in\RR^{s_c\times c}$, $\R\Sr^T\in\RR^{r\times s_r}$, and $\Sc\A\Sr^T\in\RR^{s_c\times s_r}$;
			\STATE Compute the solution of the sketched GMR as $\TX = (\Sc\C)^{\dagger}\Sc\A\Sr^T(\R\Sr^T)^{\dagger}$;
			\STATE {\bf Output:} $\TX$.
		\end{algorithmic}
\end{algorithm}

Let $\TX$ denote the solution of the sketched GMR problem~\eqref{eq:ske_gmlr}. 
One common requirement of $\TX$ is to achieve the $(1+\epsilon)$-relative error in randomized linear algebra, that is, 
\begin{equation*}
\|\A-\C\TX\R\|_{F}\leq (1+\epsilon)\min_{\X}\|\A-\C\X\R\|_{F}.
\end{equation*} 
Next, we will show what properties of the sketching matrices $\Sc$ and $\Sr$ should satisfy to guarantee $\TX$ can obtain the $(1+\epsilon)$-relative error bound.

\begin{table*}[]\setlength{\tabcolsep}{0.5 pt}
	\caption{The leverage score sampling is w.r.t.\ the column leverage scores of $\C$ and row leverage scores of $\R$, respectively. Here $\gamma$ is a positive constant. The number of the nonzero entries per column of OSNAP is $\cO(1)$.  $\rho$ is defined in Eqn.~\eqref{eq:rho}.
	}
	\label{tb:property_theo}
	\begin{center}
		\begin{tabular}{c c c c}
			\hline
			{\bf Sketching matrices}             &~~~Order of $s_c$~~~&~~~Order of $s_r$~~~&~~~$\TimeSketch$\\ \hline
			Leverage Sampling  &~~~$ \max\{\frac{c}{\sqrt{\epsilon}},  \frac{c}{\epsilon\rho^2}\}+ c\log c$~~~
			&~$ \max\{ \frac{r}{\sqrt{\epsilon}}, \frac{r}{\epsilon\rho^2}\} + r\log r  $~&~ $mc\log m+ nr\log n+s_cs_r$
			\\
			Gaussian Projection\ &~$\max\{\frac{c}{\sqrt{\epsilon}},  \frac{c}{\epsilon\rho^2}\}$~
			& ~$ \max\{ \frac{r}{\sqrt{\epsilon}}, \frac{r}{\epsilon\rho^2}\}$ ~&~$\min\{s_c,s_r\}\cdot\nnz(\A)$\\
			SRHT            &~$\max\{\frac{c}{\sqrt{\epsilon}},  \frac{c}{\epsilon\rho^2}\} + c\log c $~
			& ~$ \max\{ \frac{r}{\sqrt{\epsilon}}, \frac{r}{\epsilon\rho^2}\} + r\log r  $~&~$mn\cdot\log(\min\{s_c,s_r\})$\\
			Count Sketch &~$\max\{\frac{c}{\sqrt{\epsilon}},  \frac{c}{\epsilon\rho^2}\} + c^2$~
			& ~$ \max\{ \frac{r}{\sqrt{\epsilon}}, \frac{r}{\epsilon\rho^2}\} + r^2  $~&~$\nnz(\A)+mc+nr+\min\{ms_r,ns_c\}$\\
			OSNAP &~$\max\{\frac{c}{\sqrt{\epsilon}},  \frac{c}{\epsilon\rho^2}\} + c^{1+\gamma}$~
			& ~$ \max\{ \frac{r}{\sqrt{\epsilon}}, \frac{r}{\epsilon\rho^2}\} + r^{1+\gamma}  $~&~$\nnz(\A)+mc+nr+\min\{ms_c,ns_r\}$\\
			\hline
		\end{tabular}
	\end{center}
\end{table*}

\subsection{Error Analysis And Computation Complexity}

As we have listed in Section~\ref{subsec:sketch}, there are different kinds of sketching matrices. 
In the following theorem, we give the bound of sketching sizes of sketching matrices to keep a $(1+\epsilon)$-relative error bound and provide the detailed computational complexity.
 
\begin{theorem}\label{thm:gmp_main}
	Let $\A\in \RR^{m\times n}$, $\C\in\RR^{m\times c}$ and $\R\in \RR^{r\times n}$ be any fixed matrices with $c \ll n$ and $r\ll m$. And $0<\epsilon<1$ is the error parameter. 
	$\rho$ is defined as
	\begin{equation}
	\label{eq:rho}
	\rho 
	\triangleq  
	\frac{\norm{\A-\C\C^\dagger\A\R\R^\dagger}_F}{\norm{(\I - \C\C^{\dagger})\A\R\R^\dagger}_F + \norm{\C\C^\dagger\A(\I-\R\R^\dagger)}_F}.
	\end{equation}
	Sketching matrices $\Sc\in\RR^{s_c\times m}$ and $\Sr\in\RR^{s_r\times n}$ are described in Table~\ref{tb:property_theo}.  $\TX$ is defined as
	\begin{equation}
	\TX = (\Sc\C)^{\dagger}\Sc\A\Sr^T(\R\Sr^T)^{\dagger} \label{eq:U_hat}.
	\end{equation}
	Then we have
	\[
	\|\A-\C\TX\R\|_{F}\leq (1+\epsilon)\min_{\X}\|\A-\C\X\R\|_{F},
	\]
	with probability at least $0.95$. The time complexity of computing $\TX$ is
	\begin{equation}
	\label{eq:cc_gmr}
		\cO(s_rr^2 +s_c c^2 + s_cs_r\cdot \min(c,r)) + T_{\mathrm{sketch}},
	\end{equation}
	where $T_{\mathrm{sketch}}$ is the time cost of computing the sketches $\Sc\C$, $\R\Sr^T$, and $\Sc\A\Sr^T$.
\end{theorem}
\begin{remark}
	From Eqn.~\eqref{eq:cc_gmr}, we can observe that the computational complexity of the fast GMR consists of the cost of solving the sketched GMR and $\TimeSketch$. 
	Table~\ref{tb:property_theo} shows that the sketched GMR with Gaussian projection matrix can be solved most efficiently because the sketching sizes $s_c$ and $s_r$ required are smaller than other kinds of sketching matrices.
	However, its $T_{\mathrm{sketch}}$ is much larger.
	Hence, Gaussian projection matrix is commonly not used independently but combining with count sketch or OSNAP where after sketching by OSNAP,  Gaussian projection matrix is used to obtain a more compact sketched form.
	Furthermore, from the $\TimeSketch$ of leverage sampling. we can observe that one can solve the GMR problem approximately even the whole the matrix $\A$ is not needed to be observed. 
	Thus, the fast GMR with leverage sampling   is recommended.
\end{remark}
\begin{remark}
	From Table~\ref{tb:property_theo}, we can observe that the sketching sizes are linear to $\max\{\frac{1}{\sqrt{\epsilon}}, \frac{1}{\epsilon\rho^2}\}$. 
	If $\frac{1}{\rho^2} \leq \sqrt{\epsilon}$, then the sketching sizes only depend on $\cO\left(\frac{1}{\sqrt{\epsilon}}\right)$.
	Note that, this kind of dependence on $\epsilon$ is unknown before.
	The best dependence on $\epsilon$ is linear dependence on $\frac{1}{\epsilon}$ such as the least square regression \citep{sarlos2006improved,clarkson2013low} and low rank approximation with respect to Frobenius norm \citep{Clarkson2009Numerical}. 
	By the definition of $\rho$ in Eqn.~\eqref{eq:rho}, we know that $\rho$ is determined by $\norm{\A-\C\C^\dagger\A\R\R^\dagger}_F$ and $\norm{(\I - \C\C^{\dagger})\A\R\R^\dagger}_F + \norm{\C\C^\dagger\A(\I-\R\R^\dagger)}_F$.
	Since $\C\C^{\dagger}\A\R\R^\dagger$ is a matrix of rank $\max\{\rk(\C),\rk(\R)\}$, without loss of generality, we assume that $\C$ and $\R$ are full column and row rank, then we have $\norm{\A - \C\C^{\dagger}\A\R\R^\dagger}_F \geq \norm{\A - \A_{\min\{c,r\}}}_F$. 
	On the other hand,  $(\I - \C\C^{\dagger})\A\R\R^\dagger$ and $\C\C^\dagger\A(\I-\R\R^\dagger)$ are at most of rank $c$ and $r$, respectively.
	Thus, it holds that 
	$
	\norm{(\I - \C\C^{\dagger})\A\R\R^\dagger}_F 
	+ 
	\norm{\C\C^\dagger\A(\I-\R\R^\dagger)}_F
		\leq
	2\norm{\A_{\max\{c,r\}}}_F.
	$
	Therefore, we can upper bound $\frac{1}{\rho}$ as 
	\begin{equation*}
	\frac{1}{\rho} 
		\leq
	2\frac{\norm{\A_{\max\{c,r\}}}_F}{\norm{\A_{\setminus \min\{c,r\}}}_F}.
	\end{equation*}
	Since it holds that $c\ll n$ and $r\ll m$, $\frac{1}{\rho}$ is commonly much smaller than $\frac{1}{\epsilon^{1/4}}$.
	Thus, the sketching sizes are linear to $\frac{1}{\sqrt{\epsilon}}$.
\end{remark}

\subsection{Extensions}

In real applications, the input matrix $\A$ maybe has some special structure like symmetry or symmetric positive semi-definiteness (SPSD). The SPSD property is very common in kernel methods. In these cases, $\C = \R^T$ in Problem~\eqref{eq:glsr} and $\X^\star$ will be symmetric or SPSD. However, $\TX$ is not symmetric in Theorem~\ref{thm:gmp_main} in most cases because $\Sc$ and $\Sr$ are chosen independently. 

However, after obtaining $\TX$ in Theorem~\ref{thm:gmp_main}, with extra minor efforts, we can get a symmetric or SPSD matrix $\HX$ which also achieves a $(1+\epsilon)$-relative error bound. The key idea is based on the following proposition.
\begin{proposition}\label{fact:cvx_pro}
	Let $\mathbb{Z}\subset \RR^{m\times n}$ be a closed convex set, and suppose that $\X\in \mathbb{Z}$. For any initial approximation $\hat{\X}$, it holds that
	\[
	\|\X - \Pii_{\mathbb{Z}}(\hat{\X})\|_F\leq \|\X - \hat{\X}\|_F,
	\]
	where $\Pii_Z(\hat{\X})$ is defined as 
	\[
	\Pii_{\mathbb{Z}}(\hat{\X}) = \argmin_{\Z}\{\|\Z - \hat{\X}\|_F:\Z\in \mathbb{Z}\}.
	\]
\end{proposition}
This proposition is well-known in convex analysis. It means that the distance of two points will not increase after projecting onto a convex set. 

Let us denote the set of $n\times n$ symmetric matrices by $\HB^n$ and denote the set of $n\times n$ PSD matrices by $\HB_+^n$.
It is easy to check that $\HB^n$ and $\HB_+^n$ are convex. For a non-symmetric matrix $\X\in\RR^{n\times n}$, the projection onto $\HB^n$ is 
\begin{align}
\Pii_{\HB^n}(\X) = \frac{\X+\X^T}{2}. \label{eq:pi_H}
\end{align}

Furthermore, the projection of $\X$ onto $\HB_+^n$ relies on three steps. 
First, we project $\X$ onto $\HB^n$ and get $\X_{\sym} = \Pii_{\HB^n}(\X)$. Second, compute  an eigenvalue decomposition $\X_{\sym} = \V\D\V^T$. 
Third, form $\D_+$ by zeroing out the negative entries of $\D$. Then the projection $\X_+$ of the matrix $\X$ onto $\HB_+^n$ is
\begin{align}
\X_+ = \Pii_{\HB_+^n}(\X) = \V\D_+\V^T. \label{eq:pi_H_+}
\end{align}

\begin{table}[]\setlength{\tabcolsep}{0.5 pt}
	\caption{The leverage score sampling is w.r.t.\ the column leverage scores of $\C$. Here $\gamma$ is a positive constant. And the number of the nonzero entries per column of OSNAP is $\cO(1)$. 
	$\rho$ is defined as 
	$
	\rho = \frac{1}{2} \cdot \frac{\norm{\A - \C\C^\dagger\A\C\C^\dagger}_F}{\norm{(\I - \C\C^\dagger)\A\C\C^\dagger}_F}
	$.
	}
	\label{tb:property_sym}
	\begin{center}
		\begin{tabular}{c c c }
			\hline
			{\bf Sketching}             &~~~~Order of $s$~~~~&~~~~$\TimeSketch$~~~~\\ \hline
			Leverage Sampling  &~~~$ \max\{\frac{c}{\sqrt{\epsilon}},  \frac{c}{\epsilon\rho^2}\} + c\log c $~~~
			&~~~$ nc\log n + s^2  $~~~
			\\
			Gaussian Projection\ &~~~$ \max\{\frac{c}{\sqrt{\epsilon}},  \frac{c}{\epsilon\rho^2}\}$~~~
			& ~~~$ \nnz(\A)s  $~~~\\
			SRHT            &~~~$  \max\{\frac{c}{\sqrt{\epsilon}},  \frac{c}{\epsilon\rho^2}\} + c\log c $~~~
			& ~~~$ n^2\log s  $~~~\\
			Count Sketch &~~~$ \max\{\frac{c}{\sqrt{\epsilon}},  \frac{c}{\epsilon\rho^2}\} + c^2$~~~
			& ~~~$ \nnz(\A)  $~~~\\
			OSNAP &~~~$ \max\{\frac{c}{\sqrt{\epsilon}},  \frac{c}{\epsilon\rho^2}\} + c^{1+\gamma}$~~~
			&~~~$\nnz(\A)$\\
			\hline
		\end{tabular}
	\end{center}
\end{table}

For the GMR with symmetric or SPSD structure, the Fast GMR algorithm has the following result. 
\begin{theorem}\label{thm:sym}
	Let $\A\in\RR^{n\times n}$ be any fixed symmetric matrix, $\C$ be any $n\times c$  matrix with $c\ll n$. And $0<\epsilon<1$ is the error parameter. $\S_1\in\RR^{s\times n}$ and $\S_2\in\RR^{s\times n}$ are sketching matrices listed in Table~\ref{tb:property_sym}. Let $\TX$ be defined as
	\begin{equation}
	\label{eq:TX}
	\TX = (\S_{1}\C)^{\dagger}\S_{1}\A\S_{2}^T(\C^T\S_{2}^T)^{\dagger}.
	\end{equation}
	Then, the projection $\TX_{\sym} = \Pii_{\HB^s}(\TX)$ of the matrix $\TX$ onto $\HB^s$ satisfies
	\[
	\|\A-\C\TX_{\sym}\C^T\|_{F}\leq (1+\epsilon)\min_{\X}\|\A-\C\X\C^T\|_{F},
	\]
	with probability at least $0.95$. 
	The projection $\Pii_{\HB^s}(\cdot)$ is defined in Eqn.~\eqref{eq:pi_H}.
	
	Furthermore, if $\A$ is SPSD, then the projection $\TX_+ = \Pii_{\HB_+^s}(\TX)$ of the matrix $\TX$ onto $\HB^s_+$ satisfies
	\[
	\|\A-\C\TX_+\C^T\|_{F}\leq (1+\epsilon)\min_{\X}\|\A-\C\X\C^T\|_{F},
	\]
	with probability at least $0.95$. 
	The projection $\Pii_{\HB_+^s}(\cdot)$ is defined in Eqn.~\eqref{eq:pi_H_+}.
\end{theorem}

\begin{remark}
	The Fast GMR for the SPSD input matrix is required to conduct eigenvalue decomposition of $\TX$ defined in Eqn.~\eqref{eq:TX}.
	However, the Fast GMR still keeps computational efficiency because the $\TX$ is a $c\times c$ matrix and its eigenvalue decomposition only takes $O(c^3)$ time. 
	Therefore, the eigenvalue decomposition of $\TX$ will not bring much computational burden. 
\end{remark}

\section{Application to SPSD Matrix Approximation}

The symmetric positive semi-definite matrix approximation is an important tool in machine learning and has been widely used to speed up large-scale eigenvalue computation and kernel learning.
The famous Nystr{\"o}m method is the most popular kernel approximation method \citep{williams2001using}.
However, from the perspective of matrix approximation, the Nystr{\"o}m method is impossible to attain $(1+\epsilon)$ bound relative to $\norm{\K-\K_k}$ unless one samples $c\geq\Omega(\sqrt{nk/\epsilon})$ columns \citep{wang2013improving}.
Here $\K\in\RR^{n\times n}$ is the kernel matrix and $\K_k$ is the best rank $k$ approximation of $\K$.
This result is unsatisfying because it requires $\Omega(n^{3/2}\sqrt{k/\epsilon})$ entries of $\K$ to be computed. 

The main reason for the inefficiency of the Nystr{\"o}m method to approximate the kernel is due to the way that the core matrix $\X$ is computed. 
In fact, much higher approximation accuracy can be achieved if $\X$ is computed by minimizing the generalized matrix regression problem $\min_{\X}\norm{\K - \C\X\C^T}_F$, where $\C$ is the selected columns of $\K$. 
There have been different ways to obtain the $\C$ such as uniform random sampling and the prototype method \citep{wang2016spsd}.
By the prototype method, one can obtain such a matrix $\C\in\RR^{n\times \cO(k/\epsilon)}$ of $\K$ columns  that $\min_{\X}\norm{\K -\C\X\C^T } \leq (1+\epsilon)\norm{\K - \K_k}$.

Recently, \citet{wang15} proposed the fast positive definite matrix approximation model which aims to construct a core matrix $\X$ efficiently when the column matrix $\C\in\RR^{n\times c}$ is at hand. 
The main idea behind the fast SPSD approximation also lies in the sketching technique which constructs the core matrix $\HX$ as follows
\begin{equation}
\label{eq:fast_spd}
\HX \triangleq  (\S\C)^\dagger (\S\K\S^T)(\C^T\S^T)^\dagger,
\end{equation}
where $\S$ is a sketching matrix. 

Theorem~\ref{thm:sym} shows our Fast GMR method has the  potential to be applied  in the SPSD matrix approximation. 
In this section, we will propose an computation efficient SPSD matrix approximation method.

\subsection{Computation Efficient SPSD Matrix Approximation}

We will propose an efficient modified Nystr{\"o}m method in this section.
Similar to the Nytr{\"o}m method, we will first sample $c$ columns of the kernel matrix $\K$ uniformly and form the column matrix $\C$.
Then, we compute the leverage scores of $\C$ and apply them to form a sub-sampled intersection matrix $\S_1\K\S_2^T$ by the leverage score sampling,
where $\S_1$ and $\S_2$ are two $s\times n$ leverage score sampling matrix mutually independent.
Finally, we compute the core matrix as
\begin{equation}
\label{eq:TX_psd}
\TX_+ = \Pii_{\HB_+^s}\left((\S_1\C)^{\dagger} (\S_1\K\S_2^T) (\C^T\S_2^T)^\dagger\right),
\end{equation}
where $\Pii_{\HB_+^s}(\A)$ denotes the projection of $\A$ onto the positive semi-definite cone.
The detailed algorithm is described in Algorithm~\ref{alg:psd}.

\begin{algorithm}[tb]
	\caption{The fast SPSD Matrix Approximation.}
	\label{alg:psd}
	\begin{algorithmic}[1]
		\STATE {\bf Input:} Sample size $c$ and $s$;
		\STATE Sample $c$ columns of the kernel matrix $\K\in\RR^{n\times n}$ uniformly and compute entries of the chosen columns to form $\C$; 
		\STATE Compute the leverage scores of $\C$;
		\STATE Compute the intersection matrix $\S_1\K\S_2^T$ with leverage scores sampling matrices $\S_1$ and $\S_2$;
		\STATE Compute $\HX = (\S_1\C)^{\dagger}\S_1\K\S_2^T(\C^T\S_2^T)^{\dagger}$
		\STATE Compute the eigenvalue decomposition of $[\V, \D] = \mathrm{eig} \left( \frac{\HX+\HX^T}{2}\right)$.
		\STATE Compute $\TX_+ = \V\D_+\V^T$ where $\D_+$ is computed by zeroing out the negative entries of $\D$;
		\STATE {\bf Output:} $\C$ and $\TX_+$.
	\end{algorithmic}
\end{algorithm}

Then the output $\C$ and $\TX_+$ of Algorithm~\ref{alg:psd} have the following properties.

\begin{theorem}
	\label{thm:lw_psd}
	Let $\C\in\RR^{n\times c}$ and $\TX_+\in^{c\times c}$ be the output of Algorithm~\ref{alg:psd}. $\K\in\RR^{n\times n}$ is the kernel matrix. If the sample size $s = \cO\left( \max\{\frac{c}{\sqrt{\epsilon}},  \frac{c}{\epsilon\rho^2}\} + c\log c \right)$, where $\epsilon\in(0,1)$ and $\rho$ is defined as 
	\begin{equation}
	\label{eq:rho_K}
	\rho 
	= 
	\frac{1}{2} \cdot \frac{\|\K - \C\C^\dagger\K\C\C^\dagger\|_F}{\norm{(\I - \C\C^\dagger)\K\C\C^\dagger}_F}.
	\end{equation}
	Then it holds with probability at least $0.95$ that
	\begin{equation*}
	\|\K - \C\TX_+\C^T\|_F \leq (1+\epsilon) \min_{\X} \|\K - \C\X\C^T\|_F.
	\end{equation*}
	Moreover, the number of entries of $\K$ required to be computed is  
	\begin{equation*}
	N = nc + c^2\cdot\max\left\{\epsilon^{-1}, \; \epsilon^{-2}\rho^{-4}\right\}.
	\end{equation*}
\end{theorem}

\begin{remark}
	Theorem~\ref{thm:lw_psd} shows that to obtain a $(1+\epsilon)$-relative error rank-$k$ approximation of the kernel matrix, one can only observe a subset entries of the kernel matrix.
	Thus,  Algorithm~\ref{alg:psd} is computation efficient and can be applied in the kernel method to accelerate the speed.
\end{remark}

\subsection{Comparison With Previous Work}
\label{subsec:spsd_cmp}

\citet{wang15} has proposed the fast positive semi-definite matrix approximation  using leverage score sampling to conduct low rank kernel approximation as
\begin{equation*}
\K \approx \C\HX\C^T
\end{equation*}
where $\HX$ is computed as Eqn.~\eqref{eq:fast_spd}.
Comparing Eqn.~\eqref{eq:fast_spd} and \eqref{eq:TX_psd}, we can observe the difference lies in the sketching way.
In Eqn.~\eqref{eq:fast_spd}, it only uses a sketching matrix $\S$ to guarantee $\HX$ to be SPSD. 
In contrast, Eqn.~\eqref{eq:TX_psd} utilizes two independent leverage score sampling matrices $\S_1$ and $\S_2$ which results in the matrix 
$
(\S_1\C)^{\dagger} (\S_1\K\S_2^T) (\C^T\S_2^T)^\dagger
$
is not even symmetric.
Hence, Eqn.~\eqref{eq:TX_psd} has to project it on the SPSD matrix cone.

Comparison between Eqn.~\eqref{eq:fast_spd} and~\eqref{eq:TX_psd} shows that the core matrix constructed as Eqn.~\eqref{eq:fast_spd} is more compact and easier to implement. 
However, the method of Eqn.~\eqref{eq:fast_spd} will cause more entries of the kernel matrix required to be observed when construct the core matrix $\HX$.
The fast SPSD approximation model of \citet{wang15} requires  the observation of 
$\cO (n c^2 / \epsilon)$ entries of $\K$ to attain the $(1+\epsilon)$-relative error bound.
In contrast, Algorithm~\ref{alg:psd} only requires $\cO\left(nc + c^2\cdot\max\left\{\epsilon^{-1}, \; \epsilon^{-2}\rho^{-4}\right\}\right)$
entires of the kernel matrix to be observed.
Thus, Algorithm~\ref{alg:psd} is more attractive than the fast SPSD approximation \citep{wang15}.

\begin{table}[]\setlength{\tabcolsep}{0.5 pt}
	\caption{The leverage score sampling matrices are of size $s\times n$ w.r.t. the row leverage scores of $\C$. $\rho$ is defined in Eqn.~\eqref{eq:rho_K} with $\rho \geq \frac{1}{2}$.
	}
	\label{tb:comp_psd}
	\begin{center}
		\begin{tabular}{c c c }
			\hline
			{\bf Method}             &~~~~Order of $s$~~~~&~~~~Number of Entries to Be Observed~~~~\\ \hline
			Fast PSD \citep{wang15}  &~~~$ c\sqrt{\frac{n}{\epsilon}} $~~~
			&~~~$ nc^2\epsilon^{-1}  $~~~
			\\
			{\bf{Algorithm~\ref{alg:psd}}} &~~~$\max\{\frac{c}{\sqrt{\epsilon}},  \frac{c}{\epsilon\rho^2}\} + c\log c$~~~
			& ~~~$ nc+ c^2\cdot\max\left\{\epsilon^{-1}, \; \epsilon^{-2}\rho^{-4}\right\}  $~~~\\
			\hline
		\end{tabular}
	\end{center}
\end{table}

\section{Application In Single-Pass SVD}

In this section, we will apply the Fast GMR algorithm to the single-pass SVD algorithm and propose an efficient single pass SVD algorithm. 

\subsection{Algorithm Description}
\label{subsec:alg_desc}

To design an `efficient' single-pass SVD, we also resort to matrix sketching. Given an input matrix $\A\in\RBmn$, a target rank $k$ and an error parameter $\epsilon$, we choose two proper sketching matrices $\tilde{\Ps} = \G_R\Ps$ and $\tilde{\Ome} = \Ome^T\G_C^T$ where $\Ps\in\RR^{r_0\times m}$, $\Ome\in\RR^{c_0 \times n}$ are two OSNAP matrices with $c_0$ and $r_0$ being both of order $\cO\left(\frac{k}{\epsilon}\right)^{1+\gamma}$.
$\G_C\in\RR^{c\times c_0}$ and $\G_R\in\RR^{r \times r_0}$ are two Gaussian projection matrices with $c$ and $r$ are of order $\cO(\frac{k}{\epsilon})$. 
Using these sketching matrices,  we can get the following sketched form in $\cO(\nnz(\A))$ time
\begin{align*}
\C = \A\tilde{\Ome}
\quad\text{and}\quad
\R = \tilde{\Ps}\A.
\end{align*}  
By the properties of sketching matrices, we know that $\C$ and $\R$ preserve important structure information of column space and row space of $\A$, respectively. Let $\U_C$ and $\V_R$ denote the orthonormal bases of $\C$ and $\R^T$, respectively. We will show that
\begin{equation}
\min_{\X}\|\A - \U_C\X\V_R^T\|_F \leq (1+\epsilon) \min_{\rk (\Z) \leq k}\|\A - \Z\|_F. \label{eq:X}
\end{equation} 
We can see that the first part of above equation is a generalized matrix regression problem and its optimal solution is  
\begin{equation}
\X^\star = \U_C^T\A\V_R. \label{eq:X_star}
\end{equation}
However, the cost of computing $\X^\star$ is $\cO(\nnz(\A)k/\epsilon)$ and it needs to store the whole $\A$ in memory which contradicts with single-pass algorithm. 

To overcome the above dilemma, we resort to the Fast GMR algorithm in Section~\ref{sec:glsr}. 
Thus, we introduce another two sketching matrices $\Sc\in\RR^{s_c\times m}$ and $\Sr\in\RR^{s_r\times n}$ defined in Algorithm~\ref{alg:Pass_effi_SVD}, and then compute  
\begin{equation}
\TX = (\Sc\U_{C})^{\dagger}\underbrace{\Sc\A\Sr^T}_{\M}(\V_{R}^T\Sr^T)^{\dagger}  \label{eq:X_hat},
\end{equation} 
where $\M$ is defined in Algorithm~\ref{alg:Pass_effi_SVD}.
We can see that $\TX$ can be computed in $\cO(\nnz(\A))$ time and only single-pass of $\A$. By Theorem~\ref{thm:gmp_main}, we  can see that  $\TX$ is a good approximation of $\X^\star$ to achieve a $(1+\epsilon)$-relative error.

Finally, we compute the SVD  of $\TX$ and construct $\U$, $\Si$, and $\V$ which satisfy 
\begin{equation*}
\norm{\A - \U\Si\V^T}_F \leq (1+\epsilon) \norm{\A - \A_k}_F.
\end{equation*}

We depict the above procedure in Algorithm~\ref{alg:Pass_effi_SVD} with some modifications to obtain a single-pass SVD.

\begin{algorithm}[tb]
	\caption{Fast Single-Pass SVD.}
	\label{alg:Pass_effi_SVD}
		\begin{algorithmic}[1]
			\STATE {\bf Input:} A real matrix $\A \in \RBmn$, target rank $k$ and error parameter $\epsilon$;
			\STATE Set sketching size $r_0$, $c_0$ to be of value $\cO\left(\frac{k}{\epsilon}\right)^{1+\gamma}$, $r$ and $c$ to be of value $\cO\left(\frac{k}{\epsilon}\right)$ and $s_c$, and $s_r$ to be of value 
			$\cO\left(\max\left\{\frac{k}{\epsilon^{3/2}}, \frac{k}{\epsilon^2 \rho^2}\right\} + \left(\frac{k}{\epsilon}\right)^{1+\gamma}\right)
			$;
			\STATE Construct OSNAP matrices $\Ps\in\RR^{r_0 \times m}$,  $\Sc\in\RR^{s_c \times m}$, $\Ome\in\RR^{c_0\times n}$, $\Sr\in\RR^{s_r\times n}$ with the number of non-zero entry per column being $\cO(1)$ and Gaussian projection $\G_R \in \RR^{r\times r_0}$, $\G_C \in \RR^{c \times c_0}$.
			\STATE $\C = [\quad]$, $\R = [\quad]$, $\M = [\quad]$.
			\WHILE {$\A$ is not completely read through}
			\STATE Read next $L$-columns of columns of $\A$ denoted by $\A_L$;
			\STATE Update $\R = [\R, \G_R\Ps\A_L]$ and $\C = \C + \A_L\Ome^T\G_C^T$; 
			\STATE Update $\M = \M + \Sc\A_L\Sr$.
			\ENDWHILE
			\STATE Compute the orthonormal bases of $\C$ and $\R^T$ by $[\U_{C},\sim] = \mathrm{qr}(\C,0)$, $[\V_{R},\sim] = \mathrm{qr}(\R^T,0)$. \label{step:UV}
			\STATE Compute $\N = (\Sc\U_{C})^{\dagger}\M(\V_{R}^T\Sr^T)^{\dagger}$. \label{step:N}
			\STATE Compute the SVD decomposition $[\U_{N}, \Si, \V_{N}] = \mathrm{svd}(\N)$. 
			\STATE {\bf Output:} $\U = \U_{C}\U_{N}$, $\Si$, and $\V = \V_{R}\V_{N}$.
		\end{algorithmic}
\end{algorithm}

\subsection{Algorithm Analysis}\label{subsec:alg_analysis}

From the description of Section~\ref{subsec:alg_desc}, we can see that each step of Algorithm~\ref{alg:Pass_effi_SVD} can preserve the important information of the input matrix. Thus, Algorithm~\ref{alg:Pass_effi_SVD} can achieve a $(1+\epsilon)$-relative error just as the  following theorem states.

\begin{theorem}\label{thm:err_bnd}
	Given any matrix $\A \in\RBmn$, target rank $k$ and error parameter $\epsilon$, and  $\rho$ is defined as
	\begin{equation}
	\label{eq:rho_svd}
	\rho = \frac{\norm{\A - \U_{C}\U_{C}^T\A\V_{R}\V_{R}^T}_F}{\norm{(\I - \U_{C}\U_{C}^T)\A\V_{R}\V_{R}^T}_F+\norm{\U_{C}\U_{C}^T\A(\I - \V_{R}\V_{R}^T)}_F},
	\end{equation} 
	the outputs of Algorithm~\ref{alg:Pass_effi_SVD} satisfy that 
	\[
	\|\A - \U\Si\V^T\|_F \leq (1+\epsilon)\|\A - \A_k\|_F,
	\] 
	with probability at least $0.9$.
	Furthermore, the computational complexity of Algorithm~\ref{alg:Pass_effi_SVD} is only 
	\begin{equation*}
	\cO\left(\nnz(\A) 
	+ 
	(m+n)\left(\frac{k}{\epsilon}\right)^{2+\gamma} 
	+
	\frac{k^2}{\epsilon^2}\cdot\max\left\{\frac{k}{\epsilon^{3/2}}, \frac{k}{\epsilon^2 \rho^2}\right\}
	+ 
	\left(\frac{k}{\epsilon}\right)^{3+2\gamma}\right). 
	\end{equation*} 
	The space complexity is 
	\begin{equation*}
	\cO\left((m+n)\frac{k}{\epsilon}+ \max\left\{\frac{k^2}{\epsilon^{3}}, \frac{k^2}{\epsilon^4 \rho^4}\right\} + \left(\frac{k}{\epsilon}\right)^{2+2\gamma}\right),
	\end{equation*}
	where $0<\gamma<1$ is a constant.
\end{theorem}

From the above analysis, we can see that Algorithm~\ref{alg:Pass_effi_SVD} runs in input sparsity. And the computational complexity  is almost cubic in $\epsilon^{-1}$. This means that even when $\epsilon$ is of moderate small value, Algorithm~\ref{alg:Pass_effi_SVD} still has good performance.

It is worth noting that Algorithm~\ref{alg:Pass_effi_SVD} only requires an SVD  of an $\cO\left(\frac{k}{\epsilon}\right)\times \cO\left(\frac{k}{\epsilon}\right)$ matrix. Generally speaking, SVD  is more costly and harder to be implemented in parallel. Hence, Algorithm~\ref{alg:Pass_effi_SVD} runs very quickly even the input sizes $m$ and $n$ are large.
Furthermore, we can observe that our algorithm is memory efficient since the leading space complexity is only linear in $m{+}n$ and $\frac{k}{\epsilon}$.

\subsection{Comparison with Previous Work}\label{subsec:comparison}

Recently, \citet{tropp2017practical} proposes a single-pass SVD algorithm (described in Algorithm~\ref{alg:sp_svd_hk} in Appendix) which requires 
$\cO\left(\frac{k}{\epsilon}\cdot\nnz(\A) + m\frac{k^2}{\epsilon^2} + n \frac{k^2}{\epsilon^3}+ \frac{k^3}{\epsilon^4}\right)$ computational cost and 
$
\cO\left(m\frac{k}{\epsilon} + n\frac{k}{\epsilon^2} +\frac{k^2}{\epsilon^3}\right)
$
space cost.
We can observe that our algorithm has better performance than the one of \citet{tropp2017practical}. 
Specifically, Algorithm~\ref{alg:Pass_effi_SVD} has a much smaller sketching size $r=\cO\left(\frac{k}{\epsilon}\right)$ in contrast to the sketching size $r = \cO\left(\frac{k}{\epsilon^2}\right)$ of \citet{tropp2017practical} to achieve that
$
\|\A - \U\Si\V^T\|_F \leq (1+\epsilon)\|\A - \A_k\|_F
$
.
Furthermore, our algorithm also achieve a much lower computational cost.

In fact, our algorithm is very similar to the one of \citet{tropp2017practical}. The outputs of Algorithm~\ref{alg:Pass_effi_SVD} satisfy that
\begin{align*}
\U\Si\V^T = \underbrace{(\A\tilde{\Ome})}_{\C}\underbrace{\left((\Sc\U_C)^\dagger\Sc\A\Sr^T(\V_R^T\Sr^T)^{\dagger}\right)}_{\N}\underbrace{(\tilde{\Ps}\A)}_{\R},
\end{align*}
while the outputs of the algorithm of \citet{tropp2017practical} (Algorithm~\ref{alg:sp_svd_hk}) satisfy
\begin{align*}
\U\Si\V^T = \underbrace{(\A\tilde{\Ome})}_{\C}\underbrace{\left(\tilde{\Ps}\A\tilde{\Ome}\right)^{\dagger}}_{\N^\prime}\underbrace{(\tilde{\Ps}\A)}_{\R}.
\end{align*}
We can see that the difference between these two algorithms lies on $\N$ and $\N^\prime$. By Theorem~\ref{thm:gmp_main}, $\N$ is indeed a good approximation of $\C^\dagger\A\R^{\dagger}$ which is the optimal solution of  Problem~\eqref{eq:glsr}, while $\N^\prime$ does not have such property. Because of the way of constructing $\N^\prime$, the sketching size $r$ of $\tilde{\Ps}\in\RR^{r\times m}$ should be larger than the sketching size $c$ of $\tilde{\Ome}\in\RR^{c\times n}$. Otherwise, $\N^\prime$ will be ill-conditioned, and the performance of the algorithm of \citet{tropp2017practical} deteriorates greatly. This has been shown theoretically and empirically in \cite{tropp2017practical}.

We now  observe our comparison from the perspective of \nystrom approximation. 
If $\A$ is a kernel matrix, and $\tilde{\Ome}$ and $\tilde{\Ps}$ are the same sampling matrices, then $\N^\prime$ is just the intersection matrix and the algorithm of \citet{tropp2017practical} is just the conventional \nystrom method. 
In this case, our algorithm can be regarded as an approximate modified \nystrom method \citep{wang2013improving}. 
\citet{wang2013improving} showed that the modified \nystrom method substantially better than  the conventional \nystrom method. This also reveals the reason why our algorithm is better.
\section{Experiments}

In this section, we will study the algorithms proposed in previous sections  empirically.
In Section~\ref{subsec:GMR}, we conduct experiments on different real-world matrices to validate the tightest of our theoretical analysis in Section~\ref{sec:glsr}.
In Section~\ref{subsec:kernel}, we will apply our fast GMR based SPSD approximation to the kernel approximation and compare it with several conventional kernel approximation methods.
In Section~\ref{subsec:svd}, we compare our fast single pass SVD algorithm with  the practical single pass SVD proposed by \citet{tropp2017practical}.

\begin{table*}[]
	\centering
	\caption{Datasets summary(sparsity$=\frac{\#\text{Non-Zero Entries}}{n\times d}$)}
	\label{tb:data}
	\begin{tabular}{ccccc}
		\hline
		Dataset~~~~ &~~~~ $m$~~~~&~~~~$n$~~~~&~~~~sparsity~~~~&~~~~source \\ \hline
		gisette~~~~ &~~~~ $5,000$~~~~&~~~~$6,000$~~~~&~~~~dense~~~~&~~~~libsvm dataset   \\
		mnist~~~~&~~~~$60,000$~~~~&~~~~$780$~~~~&~~~~dense~~~~&~~~~libsvm dataset    \\
		svhn~~~~    &~~~~  $19,082$~~~~&~~~~$3,072$~~~~&~~~~dense~~~~&~~~~libsvm dataset     \\ 
		rcv1~~~~&~~~~$20,242$~~~~ &~~~~ $50,236$~~~~ &~~~~$0.16\%$~~~~ &~~~~libsvm dataset     \\ 
		real-sim ~~~~&~~~~$72,309$~~~~&~~~~$20,958$~~~~&~~~~$0.24\%$~~~~&~~~~libsvm dataset      \\ 
		news20~~~~&~~~~$15,935$~~~~&~~~~$62,061$~~~~&~~~~$0.13\%$~~~~&~~~~libsvm dataset     \\
		\hline
	\end{tabular}
\end{table*}
\subsection{Experiments on Fast GMR}
\label{subsec:GMR}
In Section~\ref{sec:glsr}, we have provided a tight bound for sketched generalized matrix approximation theoretically. 
It is of interest to empirically validate the correctness and tightness of our theoretical analysis. 

\begin{figure}[!ht]
	\subfigtopskip = 0pt
	\begin{center}
		\centering
		\subfigure[\textsf{gisette}]{\includegraphics[width=55mm]{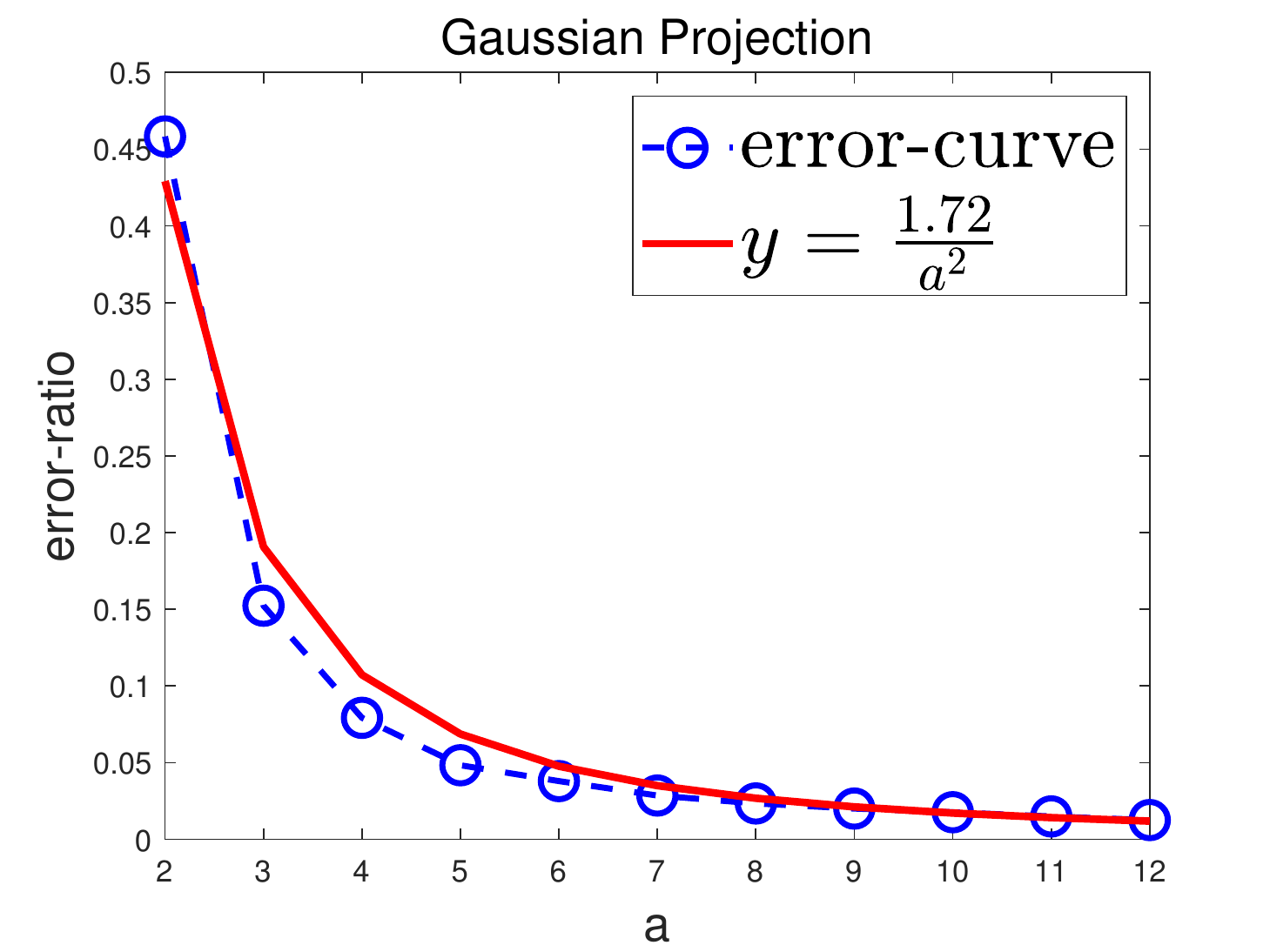}}~
		\subfigure[\textsf{mnist}]{\includegraphics[width=55mm]{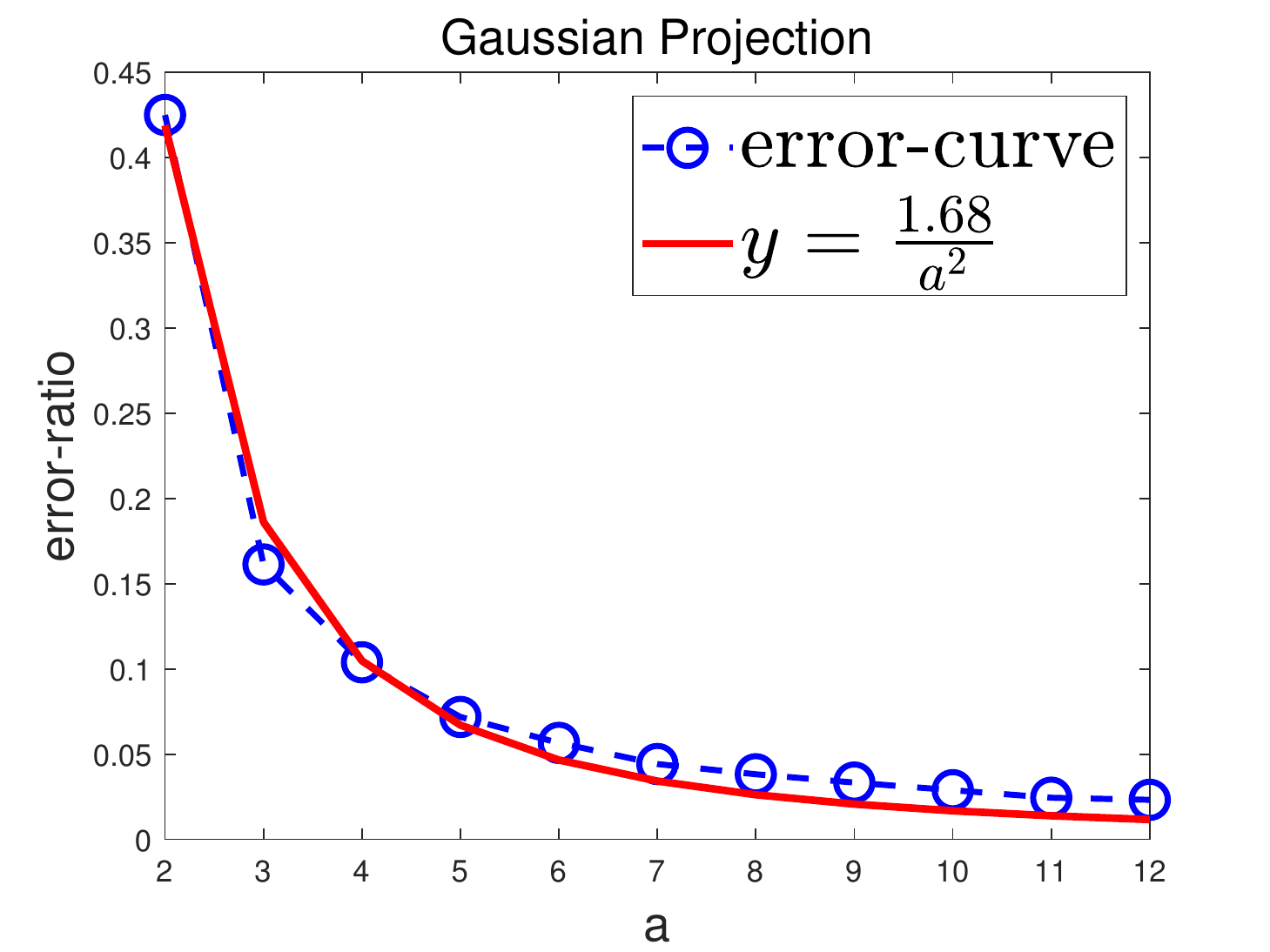}}~
		\subfigure[\textsf{madelon}]{\includegraphics[width=55mm]{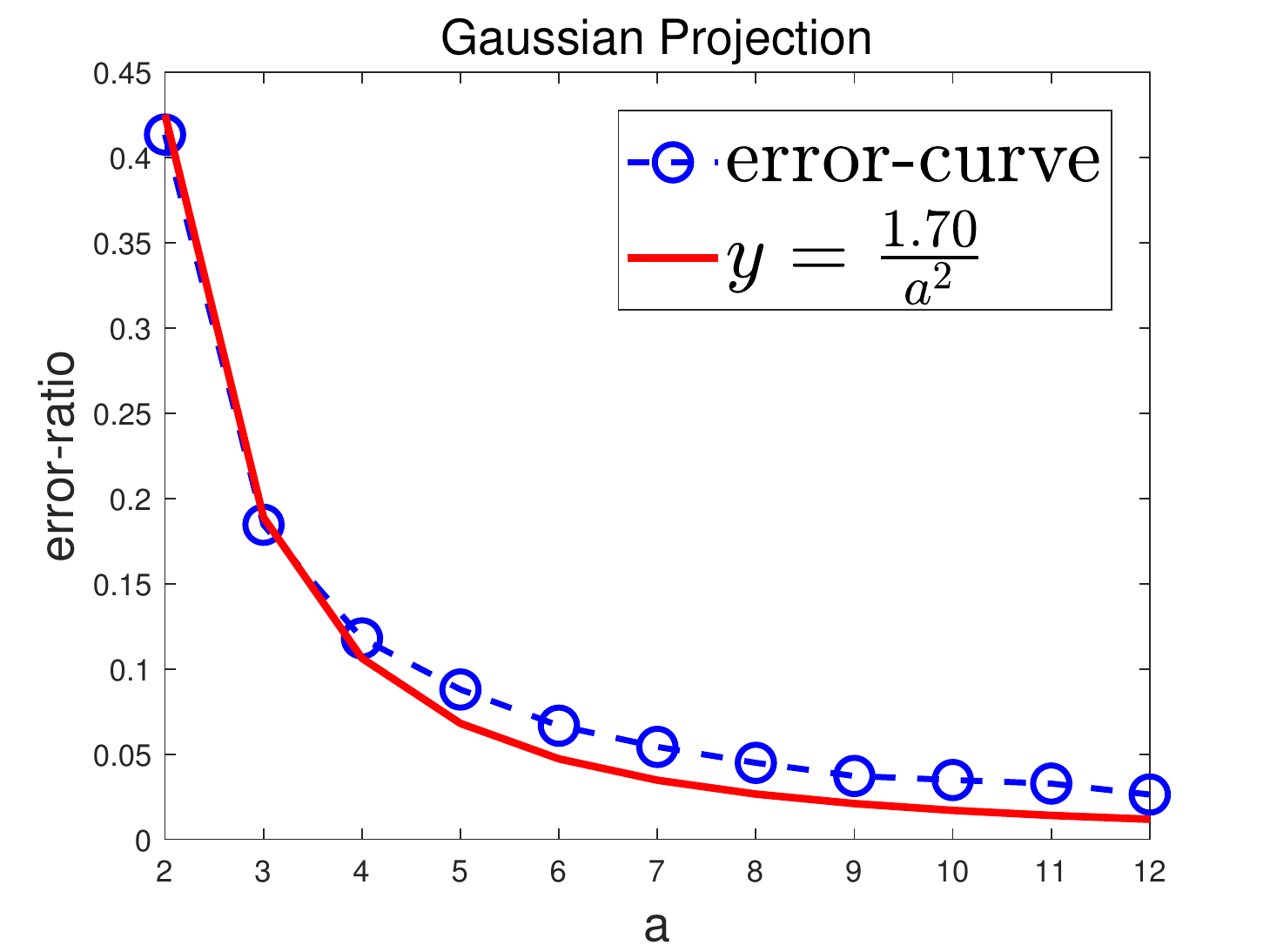}}~\\
		\subfigure[\textsf{rcv1}]{\includegraphics[width=55mm]{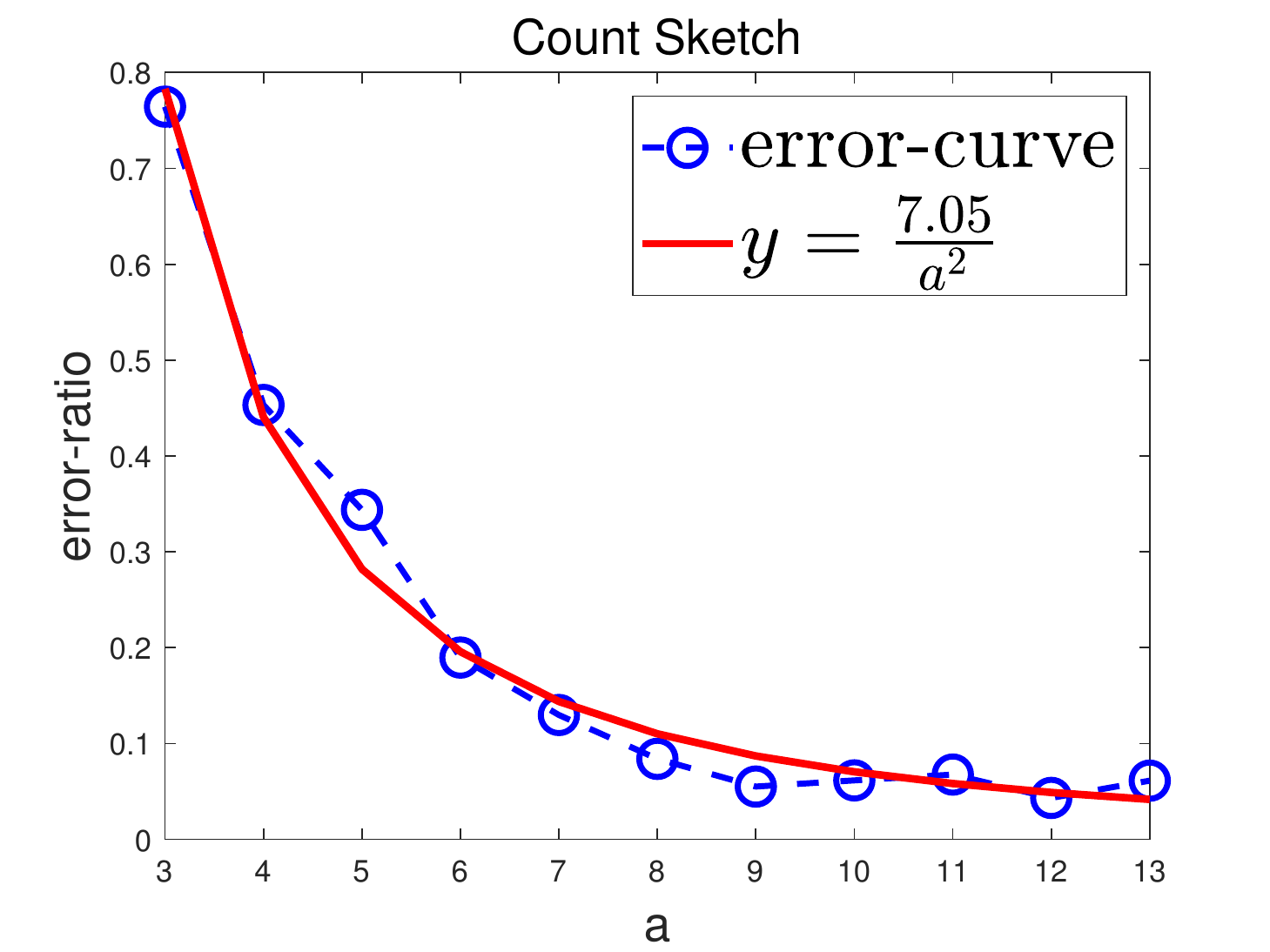}}~
		\subfigure[\textsf{real-sim}]{\includegraphics[width=55mm]{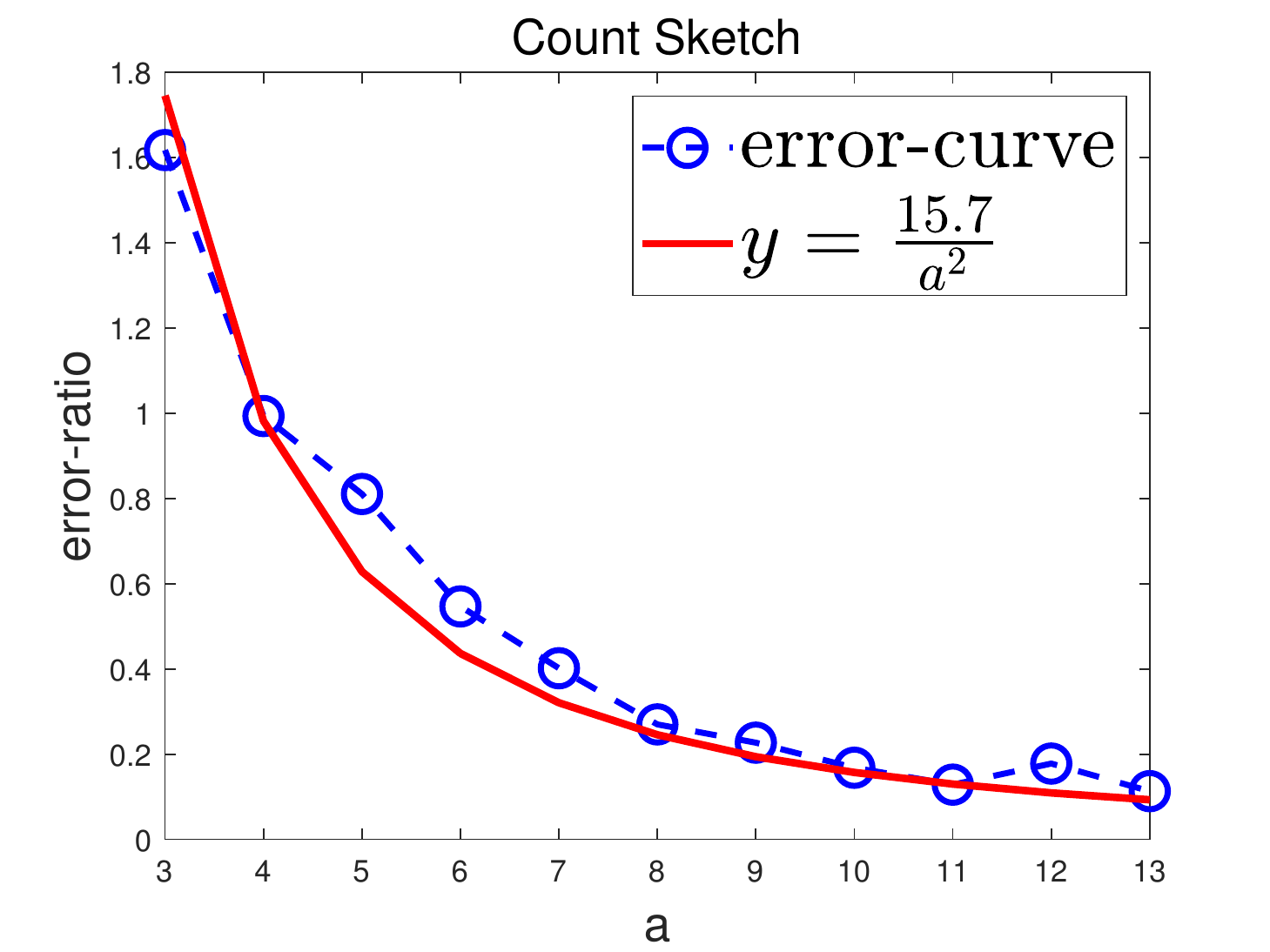}}~
		\subfigure[\textsf{news20}]{\includegraphics[width=55mm]{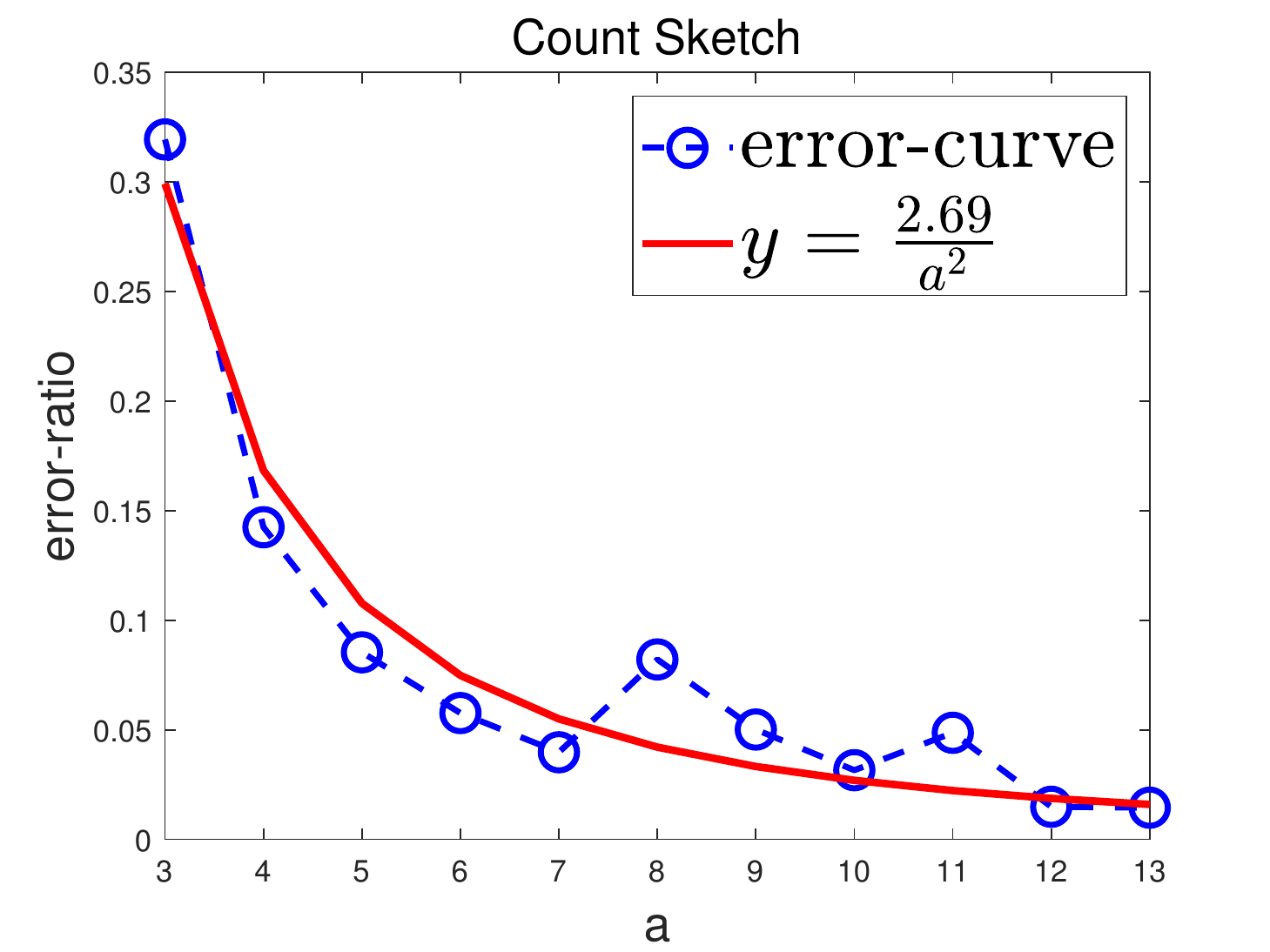}}
	\end{center}
	\caption{Result of Comparison}
	\label{fig:glsr}
\end{figure}

We conduct several experiments on real-world datasets whose detailed description  is listed in Table~\ref{tb:data}. 
For the $\C$ and $\R$, we construct them as
\begin{equation*}
\C = \A\G_C \quad\mbox{and}\quad \R = \G_R\A,
\end{equation*} 
where $\G_C\in\RR^{n\times c}$ and $\G_R\in\RR^{r\times m}$ are two Gaussian projection matrices.
In our experiments, we set $c = 20$ and $r = 20$.
In the following experiments, we choose our sketching matrices $\Sc$ and $\Sr$ as Gaussian projection for dense matrices and count sketch matrices for sparse matrices, respectively. 
In addition, we set $s_c = ac$ and $s_r = ar$ where $a$ varies from $2$ to $12$ for dense matrix $\A$ and $a$ varies from $3$ to $13$ for sparse matrix $\A$. We report the error ratio is defined as
\[
\text{error ratio} = \frac{\|\A-\C\TX\R\|_F}{\|\A-\C\C^\dagger\A\R^\dagger\R\|_F} - 1,
\] 
where $\TX$ is defined in Eqn~\eqref{eq:U_hat}. 
Furthermore, to compute $\|\A-\C\TX\R\|_F$ efficient when $\A$ is a large sparse matrix, we will use the following method to approximate it
\begin{equation*}
\|\S_1(\A-\C\TX\R)\S_2\|_F = (1\pm \epsilon) \|\A-\C\TX\R\|_F,
\end{equation*} 
where $\S_1\in\RR^{s_1\times m}$ and $\S_2\in\RR^{n\times s_2}$ are two count sketch matrices with $s_1,s_2 = \cO(\epsilon^{-2})$ \citep{clarkson2013low}.
$\|\A-\C\C^\dagger\A\R^\dagger\R\|_F$ can be approximated similarly efficiently.
We report our result in Figure~\ref{fig:glsr}.

We can observe that the error ratio is linear to $1/a^2$ which shows that the sketching sizes $s_c$ and $s_r$ is of order $\cO\left(\frac{c}{\sqrt{\epsilon}}\right)$ and $\cO\left(\frac{r}{\sqrt{\epsilon}}\right)$, respectively.
This validates the tightness of our sketching sizes bound shown in Theorem~\ref{thm:gmp_main}. 
Furthermore, we can also observe that when $a = 10$, the error ratio $\epsilon$ is really small and is close to $0.05$ on most datasets.
Thus, our Fast GMR algorithm can achieve good performance in real applications with sketching sizes only being of several times of $c$ and $r$. 

\begin{table*}[]
	\centering
	\caption{Datasets summary for kernel approximation}
	\label{tb:ker_data}
	\begin{tabular}{ccccccc}
		\hline
		Dataset 	& dna 	& gisette      	& madelon & mushrooms & splice & a5a \\
		\hline
		\#Instance 	& 2,000 & 6,000 		& 2,000 	&8,142	  & 1,000	&6,414\\
		\#Attribute	& 180 	& 5,000			& 500 		&112	  & 60	&123\\
		$\sigma$	& 0.04	& $1.5\times 10^{-3}$	& $3.5\times 10^{-6}$ & 0.1 & 0.02  & 0.3\\
		$\eta$		& 0.89	& 0.85			& 0.87		& 0.95 		& 0.83	& 0.63\\
		\hline
	\end{tabular}
\end{table*}

\subsection{Experiments on Kernel Approximation}
\label{subsec:kernel}

Let $X = [\x_1,\dots,\x_n]$ be the $d\times n$ data matrix, and $\K$ be the RBF kernel matrix with each entry computed by $\K_{i,j} = \exp\left(-\sigma\|\x_i - x_j\|^2\right)$,
where $\sigma$ is the scaling parameter.
We set the scaling parameter $\sigma$ in the following way. 
We fix $k = 15$ and define
\begin{equation*}
\eta = \frac{\|\K_k\|_F^2}{\|\K\|_F^2} = \frac{\sum_{i=1}^{k}\lambda^2_i(\K)}{\sum_{i=1}^{n}\lambda^2_i(\K)}.
\end{equation*}
We choose $\sigma$ such that $\eta$ is above $0.6$. 
We conduct experiments on several datasets available at the LIBSVM site. The datasets are summarized in Table \ref{tb:ker_data}. 

\begin{figure}[!ht]
	\subfigtopskip = 0pt
	\begin{center}
		\centering
		\subfigure[\textsf{dna}]{\includegraphics[width=55mm]{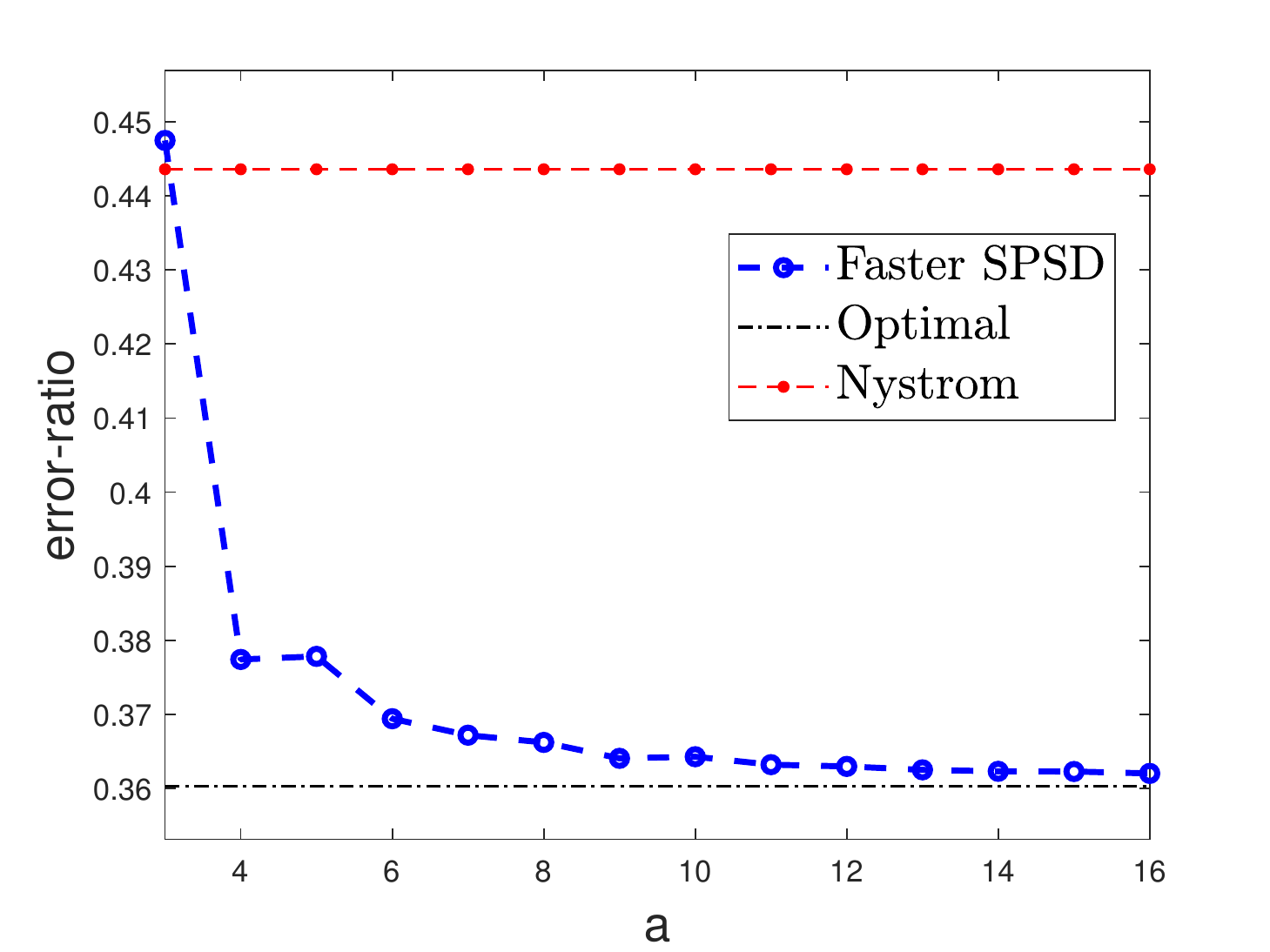}}~
		\subfigure[\textsf{gisette}]{\includegraphics[width=55mm]{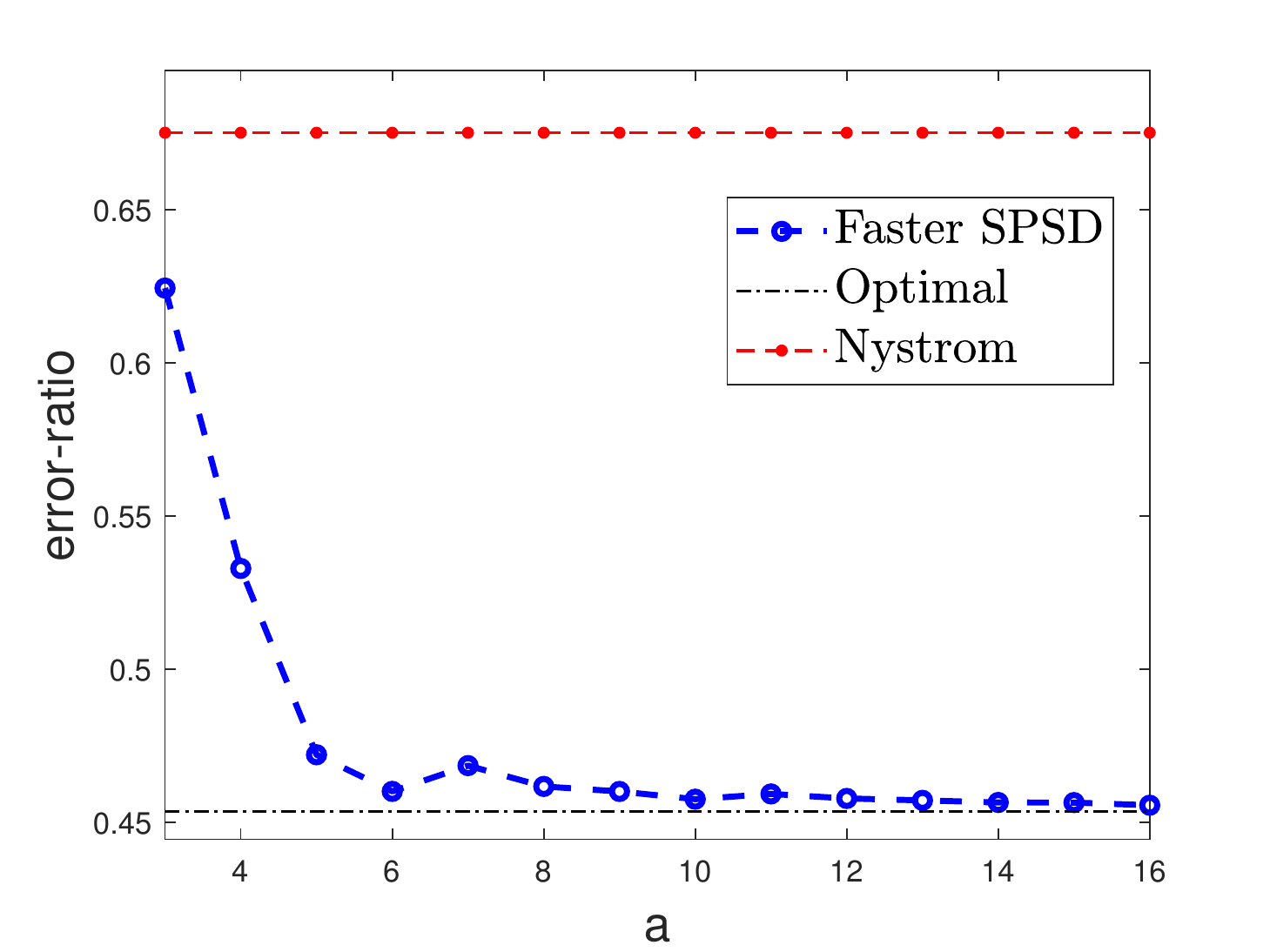}}~
		\subfigure[\textsf{madelon}]{\includegraphics[width=55mm]{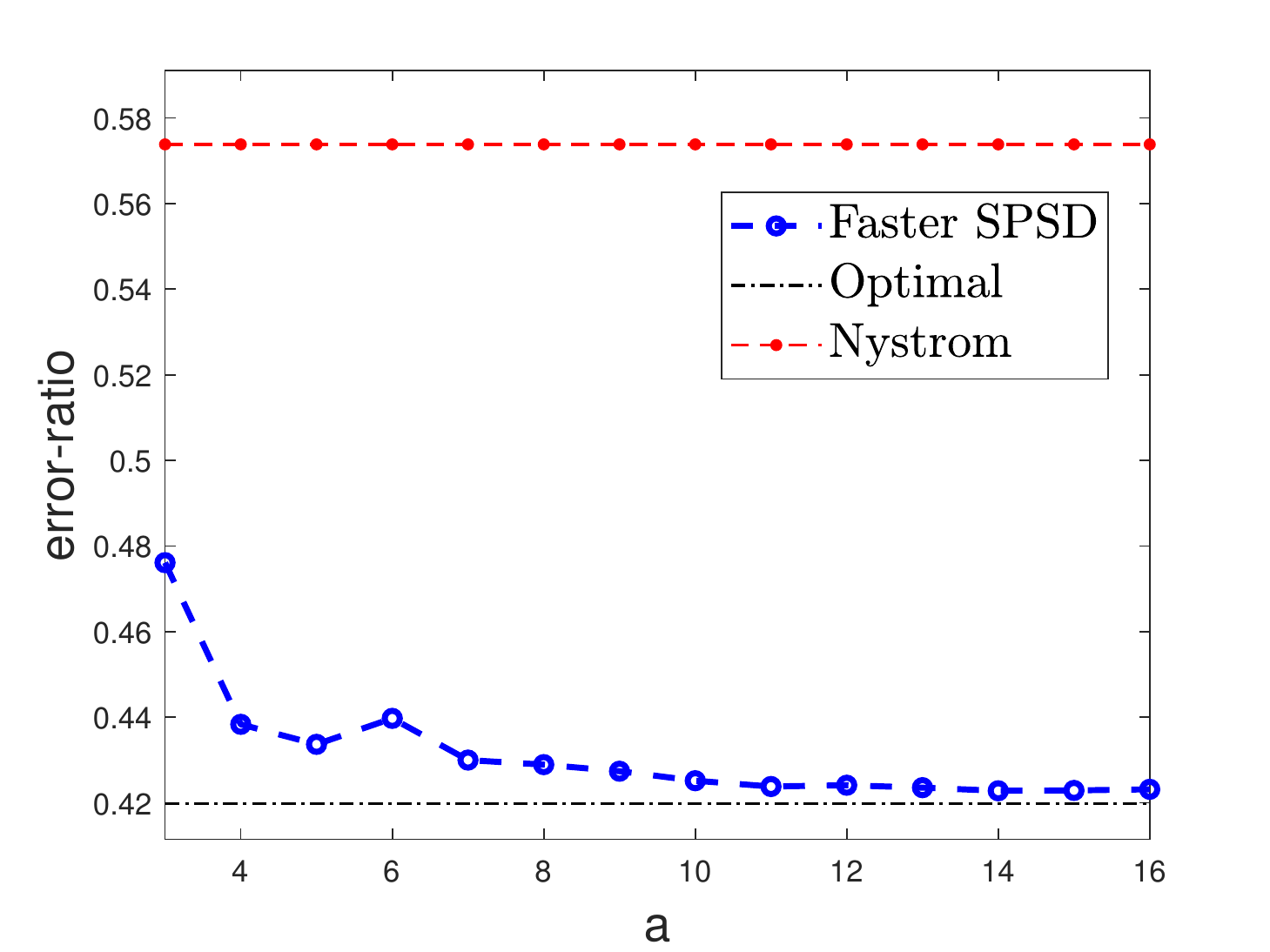}}~\\
		\subfigure[\textsf{mushrooms}]{\includegraphics[width=55mm]{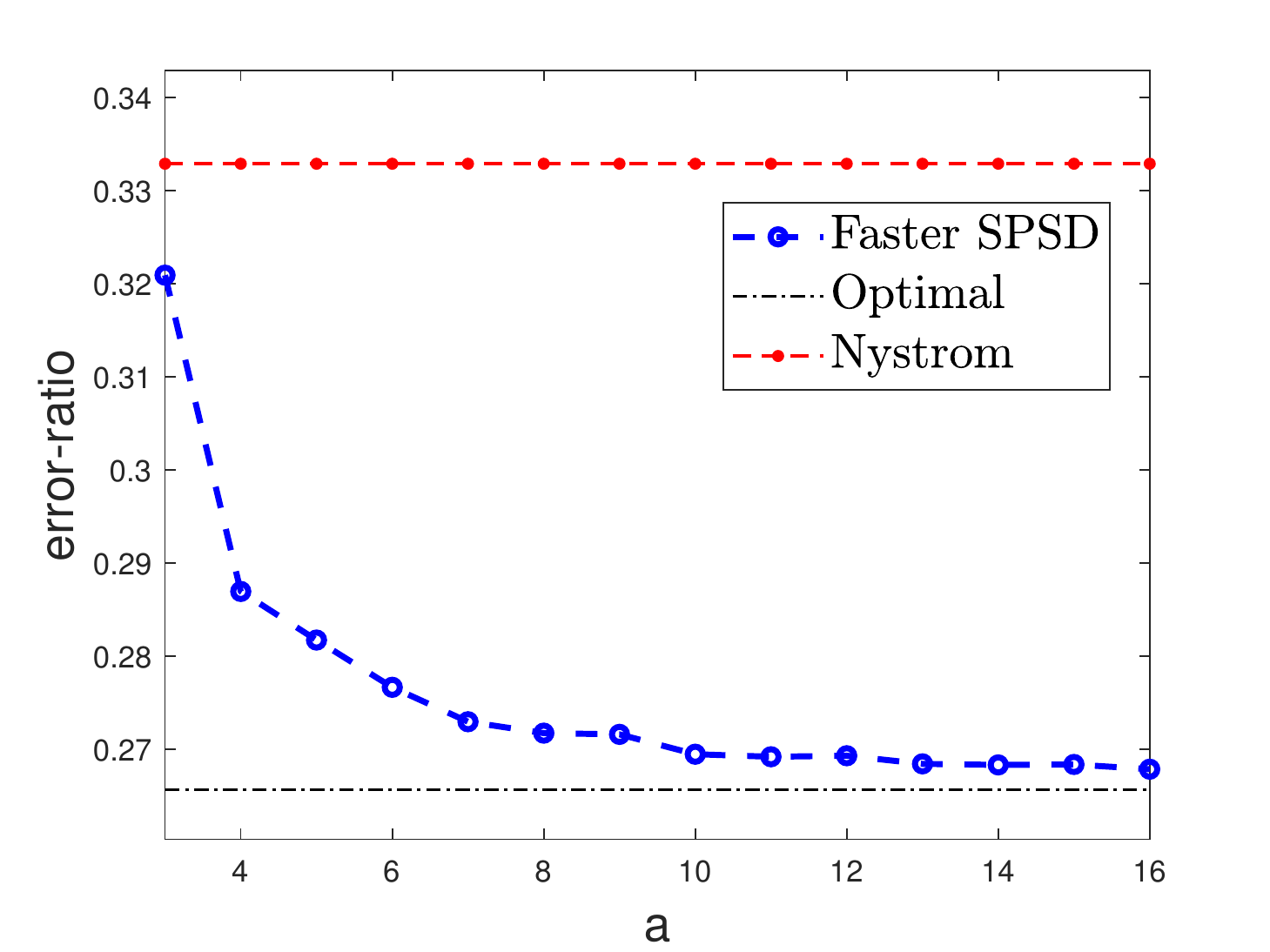}}~
		\subfigure[\textsf{splice}]{\includegraphics[width=55mm]{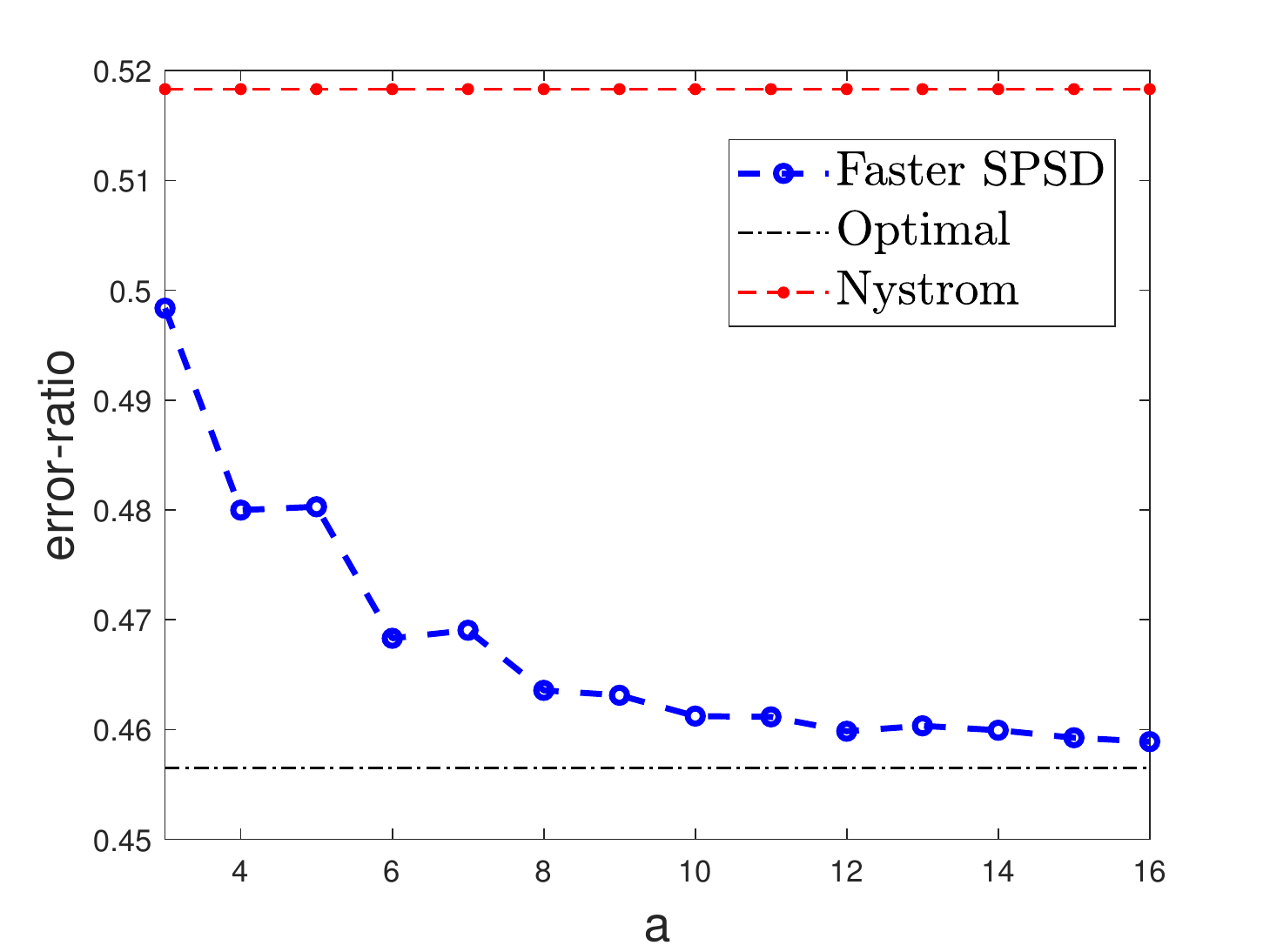}}~
		\subfigure[\textsf{a5a}]{\includegraphics[width=55mm]{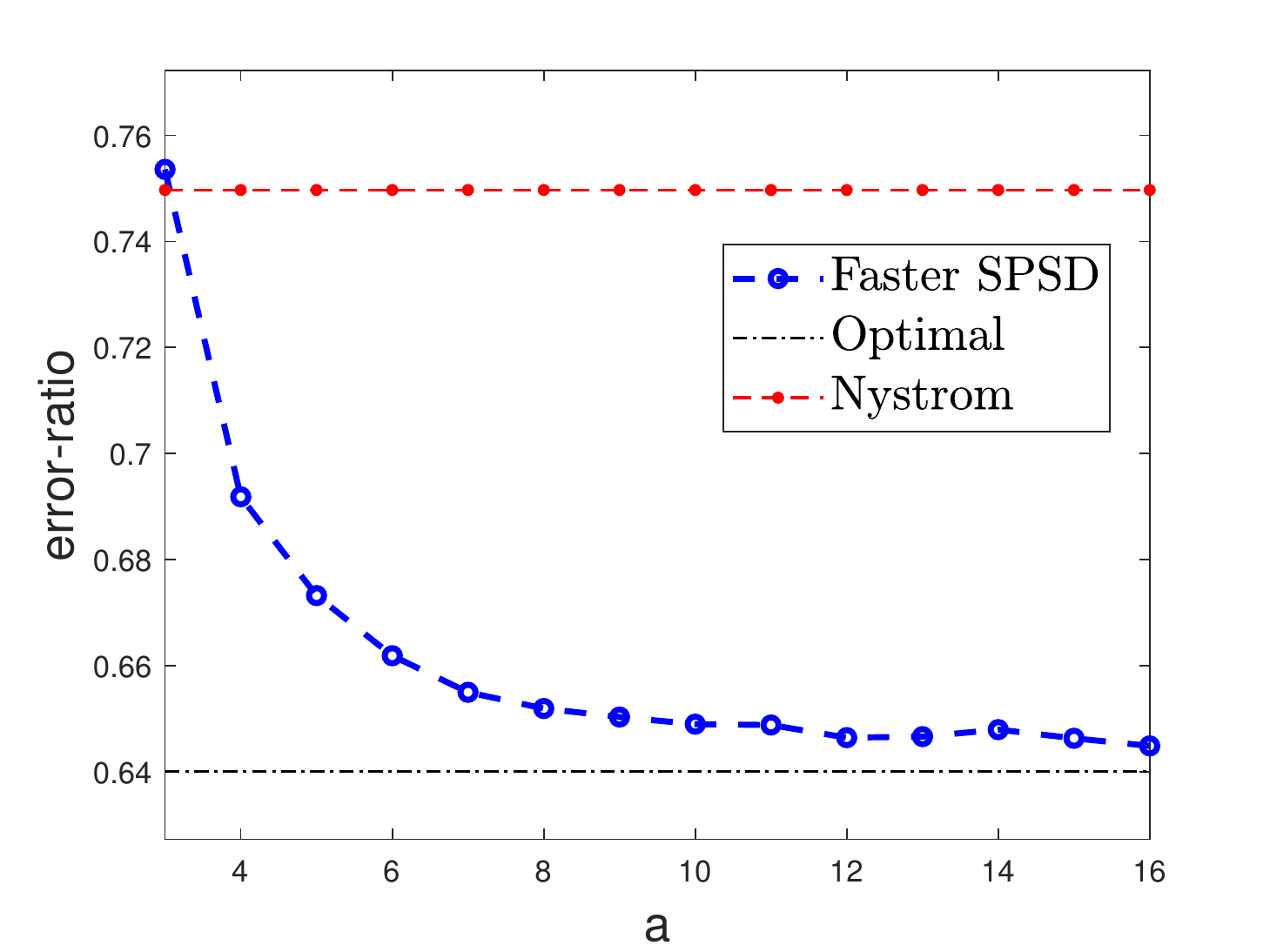}}
	\end{center}
	\caption{Result of Comparison}
	\label{fig:fast_spsd}
\end{figure}

In this set of experiments, we study the effect of how the core matrix $\X$ is constructed. We form $\C\in\RR^{n\times c}$ by uniform sampling with $c = 2k$. 
To evaluate the sketching size $s$ affecting the approximation accuracy, we choose $s = ac$ with $a$ varying from $3$ to $16$ and plot $\frac{s}{c}$ against the approximation error 
$$\text{error ratio} = \frac{\|\K - \C\X\C^T\|_F}{\|\K\|_F}.$$ 
We compare our algorithm with \nystrom, the fast SPSD algorithm \citep{wang15}, and the optimal method which constructs the optimal core matrix $\X = \C^\dagger\K(\C^\dagger)^T$.
To distinguish with the the fast SPSD algorithm \citep{wang15}, we name our method (Algorithm~\ref{alg:psd}) faster SPSD method.
We report experiments result in Figure~\ref{fig:fast_spsd} and list the result of the fast SPSD in Table~\ref{tb:spsd}.

From the Figure~\ref{fig:fast_spsd}, we can observe that once $s = 10c$, our faster SPSD algorithm can achieve almost good as the optimal error ratio.
In contrast, there is a large gap between \nystrom and the optimal error ratio.
Then, we compare our faster SPSD algorithm with the fast SPSD method whose result is listed in Table~\ref{tb:spsd}.
We can observe that when $s$ is of several times of $a$, the fast SPSD perform poorly and the error ratio is commonly much larger than the one of \nystrom method. 
For example, on the kernel matrix based on `madelon' dataset, the error ratio of \nystrom is only about $0.55$ while the error ratio of the fast SPSD is $1.60$ even $s/c = 16$.
Instead, our faster SPSD achieve a error ratio close to $0.42$.
In fact, the fast  SPSD can achieve better error ratio than \nystrom and close to the optimal ratio only when $s$ is one tenth of $n$ just shown in the experiments of \citet{wang15}.
Furthermore, the experiments also validate theoretical analysis in Section~\ref{subsec:spsd_cmp}.
Therefore, we claim that our faster SPSD algorithm outperforms the fast SPSD both theoretically and empirically. 

\begin{table}
	\centering
	\caption{The error ratio against of $s/c$ of the fast SPSD \citep{wang15}}
	\label{tb:spsd}
	\begin{tabular}{c|cccccc}
		\hline
		\diagbox[width=7em]{$a = s/c$}{dataset} &~ dna & gisette & madelon & mushrooms & splice & a5a~~~\\
		\hline
		$a = 8$  &~ 1.06 & 12.8 & 3.07 & 0.44	& 1.26	& 1.83~~~\\
		$a = 10$ &~ 0.95 & 9.83 & 2.44 &	0.43	& 1.04 	& 1.49~~~\\
		$a = 12$ &~ 0.78 & 8.47 & 2.04 &	0.39	& 0.90	& 1.34~~~\\
		$a = 14$ &~ 0.72 & 7.08 & 1.79 &	0.30	& 0.80	& 1.12~~~\\
		$a = 16$ &~ 0.66 & 6.03 & 1.60 &	0.33	& 0.73	& 1.01~~~\\
		\hline
	\end{tabular} 
\end{table}

\subsection{Experiments on Single Pass SVD }
\label{subsec:svd}

We fix the input matrix $\A\in\RBmn$ and the target rank $k$. 
Let $\U$, $\Si$, and $\V$ be the outputs of the single pass SVD algorithm.
We will measure the relative error to the best rank $k$ approximation:
\begin{equation}
\label{eq:err_ratio_svd}
\text{error ratio} = \frac{\|\A - \U\Si\V^T\|_F}{\|\A -\A_k\|_F} - 1.
\end{equation} 
We compare Algorithm~\ref{alg:Pass_effi_SVD} (Fast SP-SVD) with the practical single pass SVD (Practical SP-SVD) algorithm of \citet{tropp2017practical}. 
This is because the existing single pass SVD algorithms almost share the same algebraic form while Practical SP-SVD implemented in a more numerical stable way just shown in the experiments of \citet{tropp2017practical}. 
It is worthy to notice both the `Fast SP-SVD' and  `Practical SP-SVD' are without fixed rank, that is, $\U$, $\Si$, and $\V$ are of rank larger than $k$. 
Thus, the relative error defined in Eqn.~\eqref{eq:err_ratio_svd} can be negative and lower bounded by $-1$.

\begin{figure}[!ht]
	\subfigtopskip = 0pt
	\begin{center}
		\centering
		\subfigure[\textsf{gisette}]{\includegraphics[width=55mm]{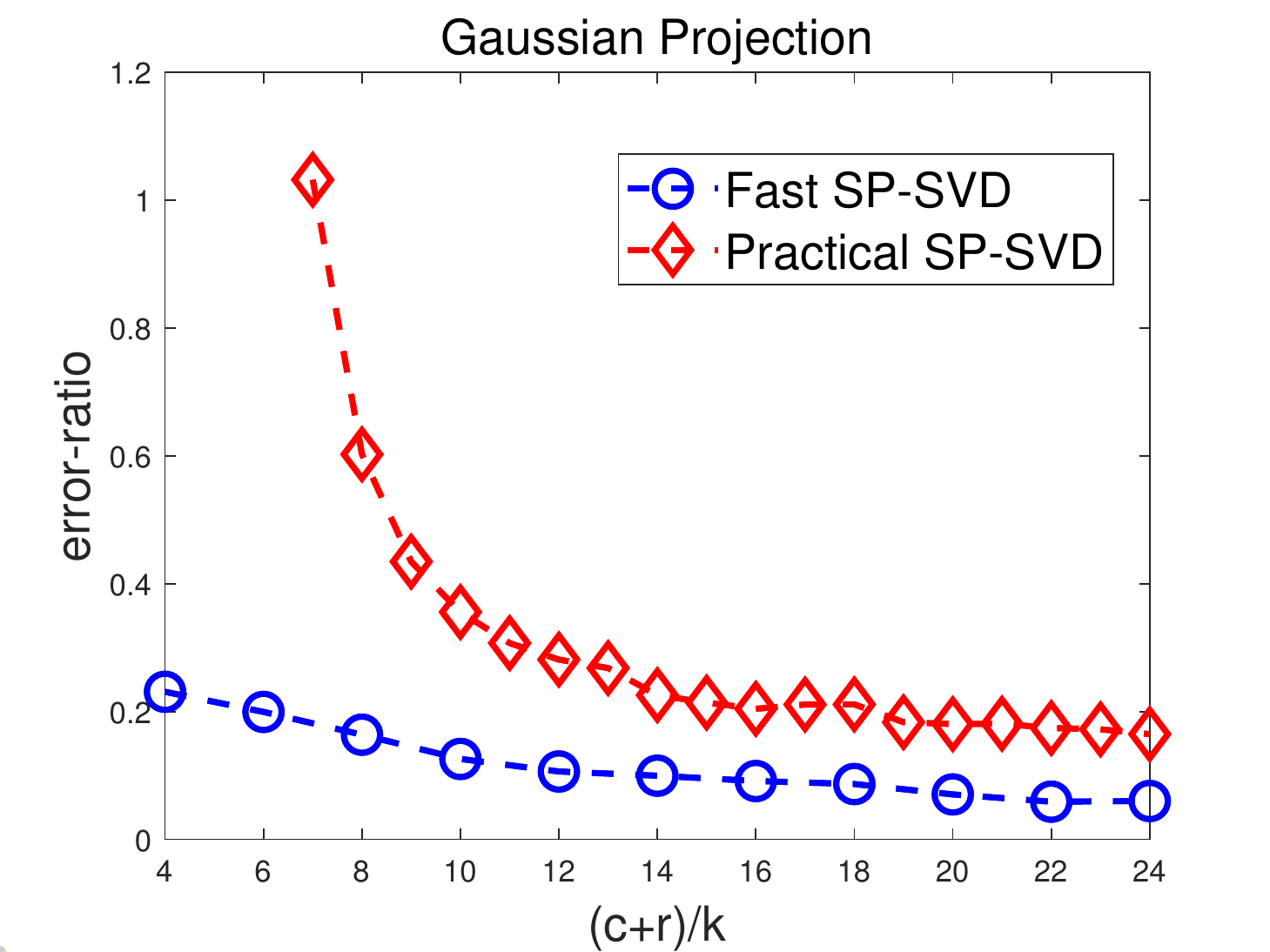}}~
		\subfigure[\textsf{mnist}]{\includegraphics[width=55mm]{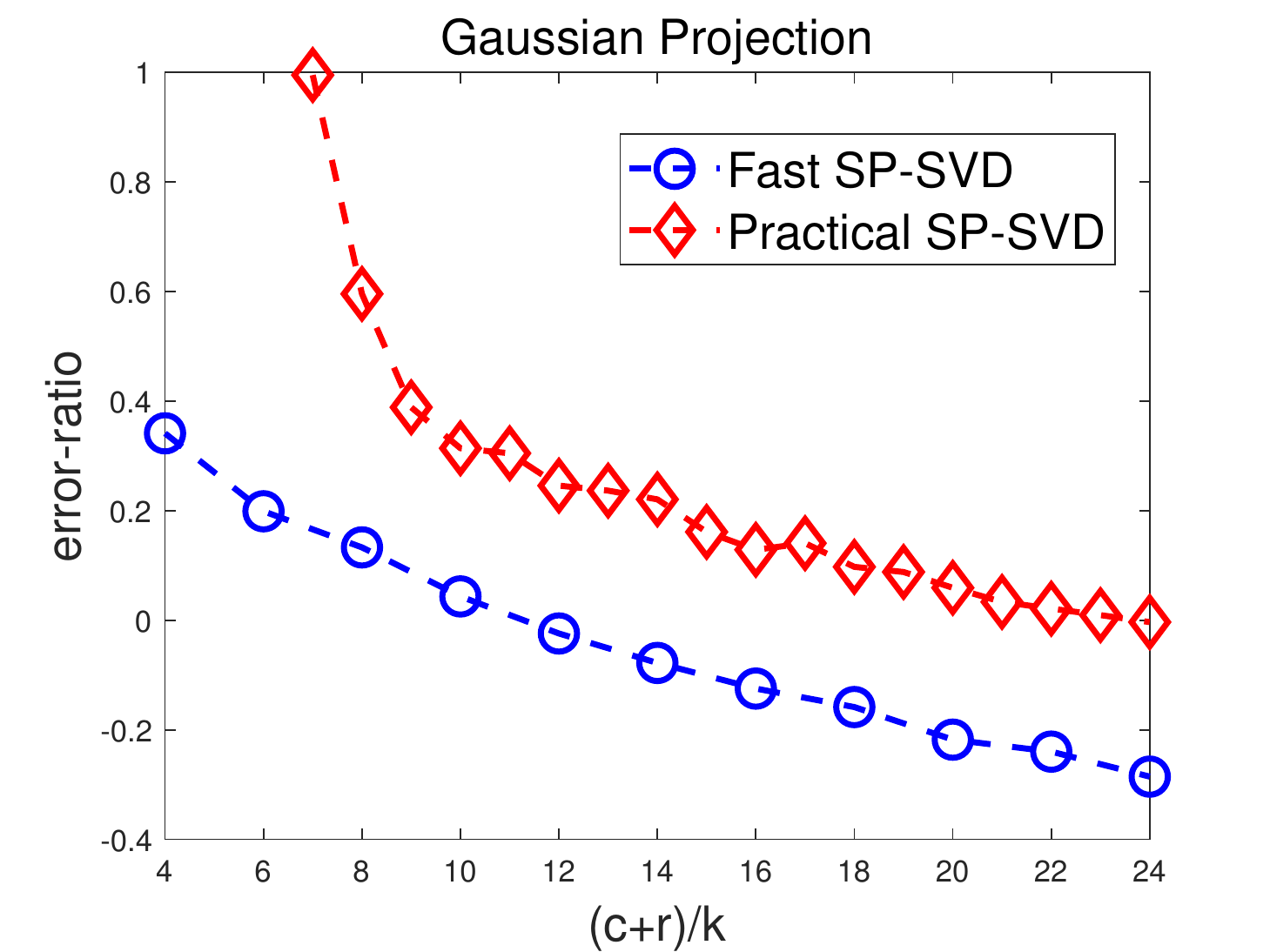}}~
		\subfigure[\textsf{madelon}]{\includegraphics[width=55mm]{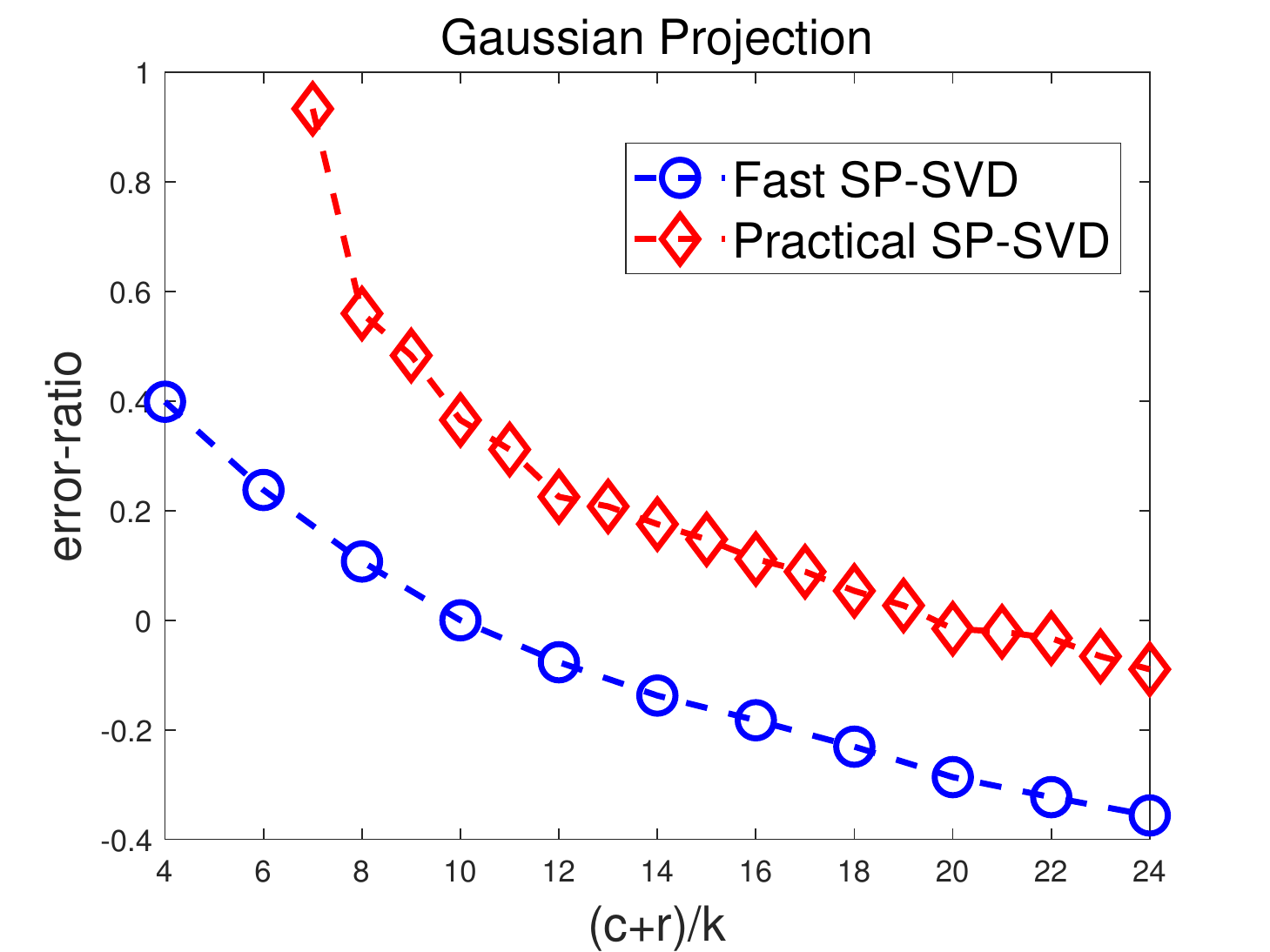}}~\\
		\subfigure[\textsf{rcv1}]{\includegraphics[width=55mm]{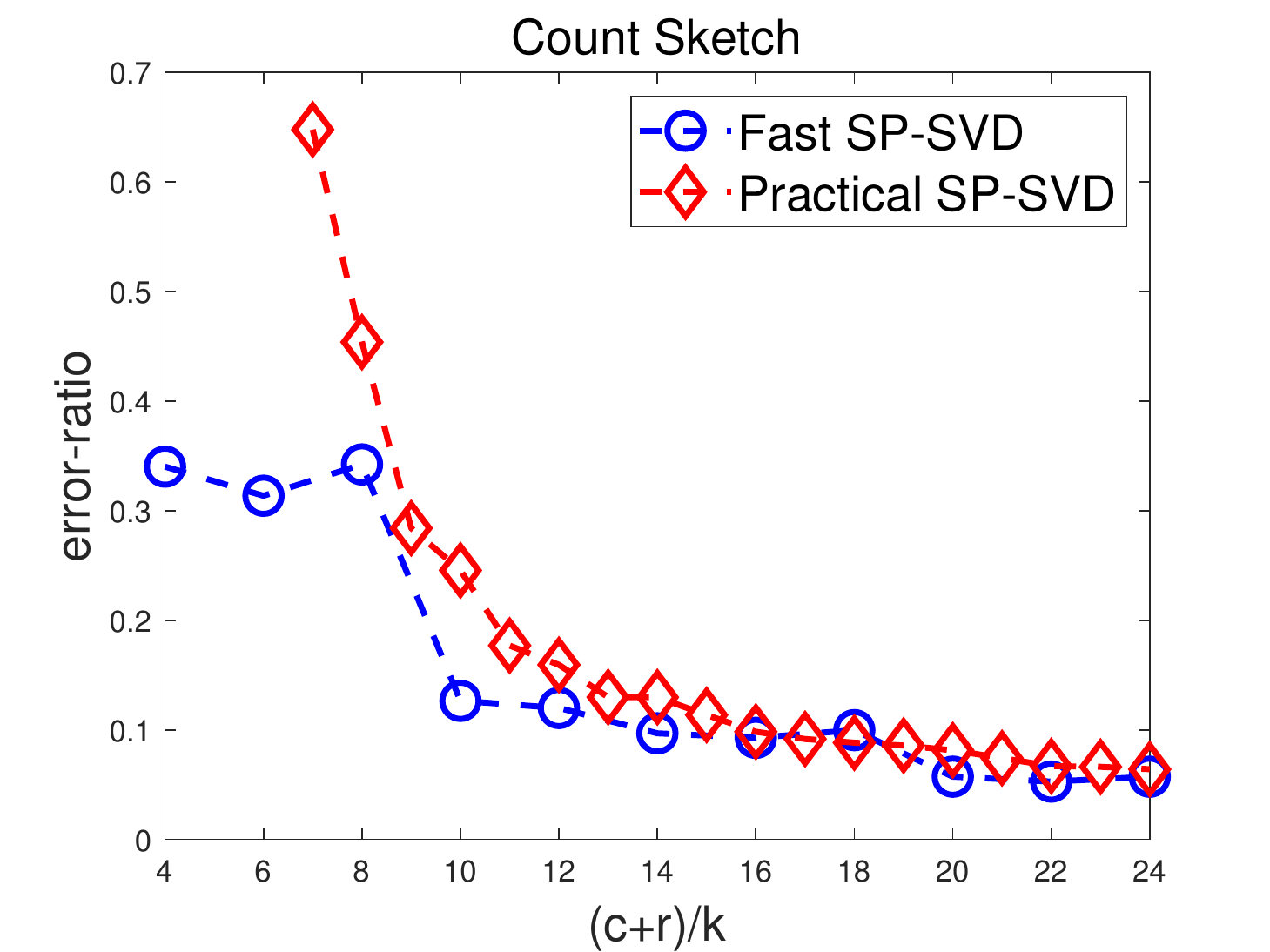}}~
		\subfigure[\textsf{real-sim}]{\includegraphics[width=55mm]{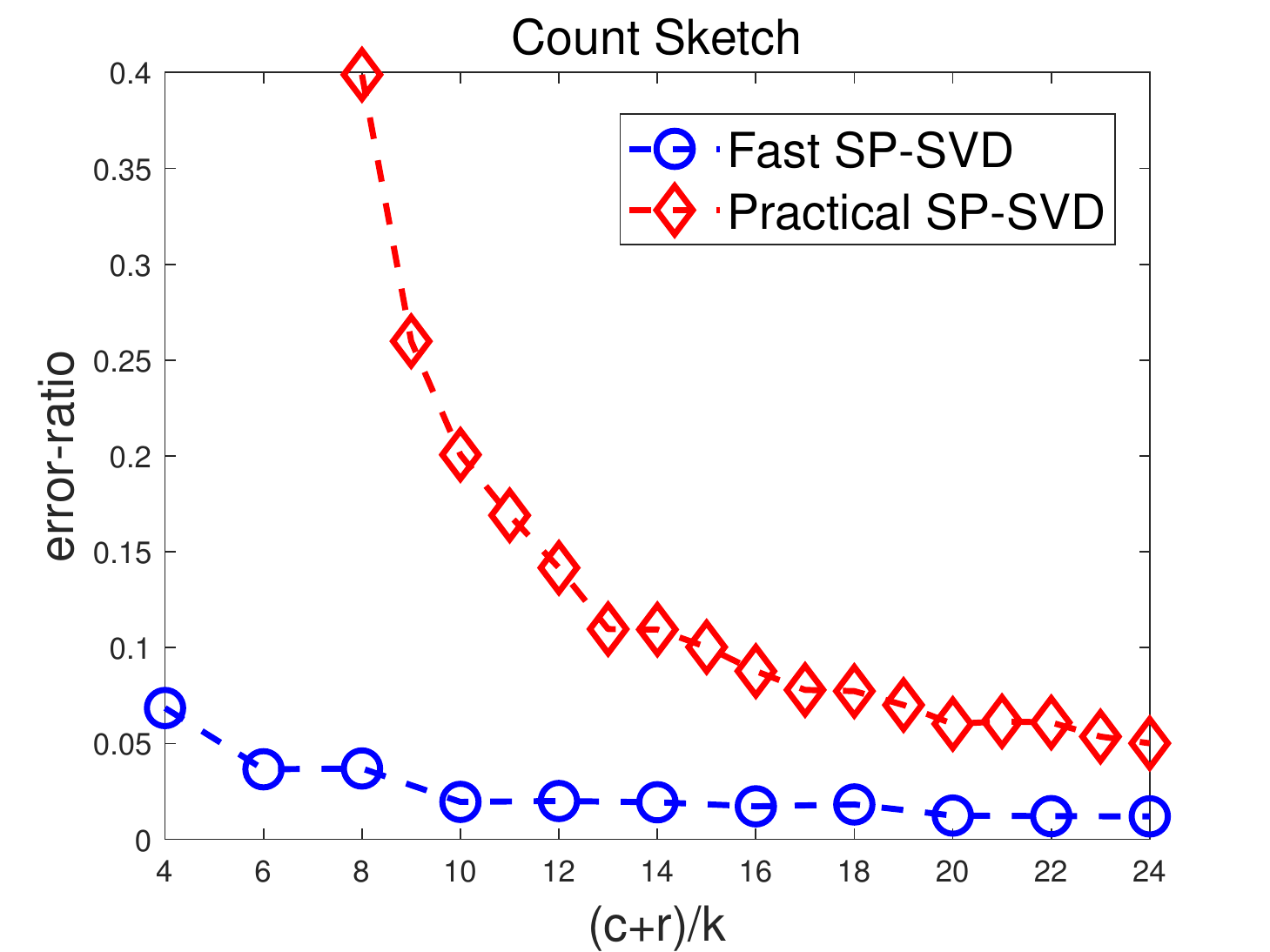}}~
		\subfigure[\textsf{news20}]{\includegraphics[width=55mm]{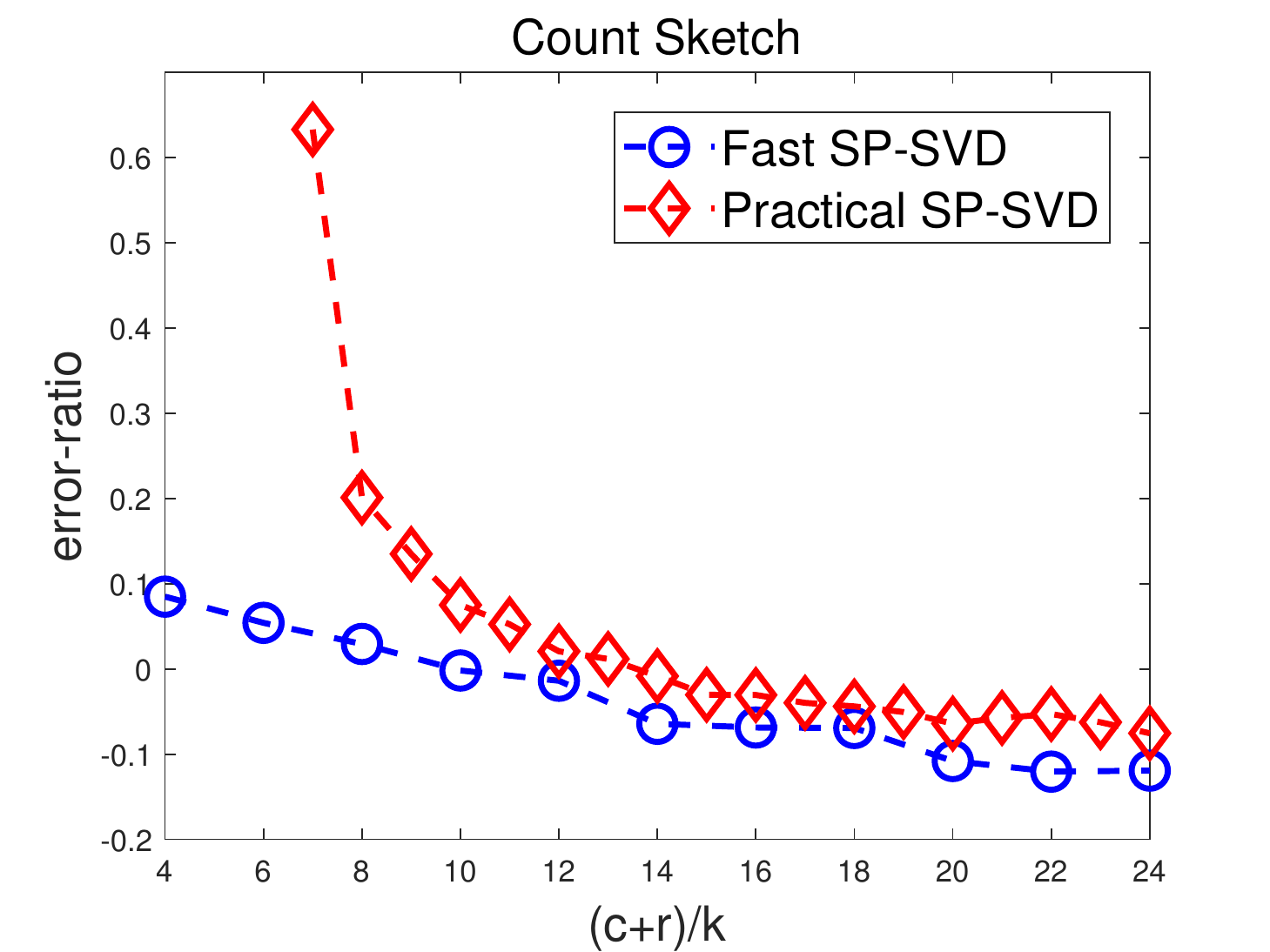}}
	\end{center}
	\caption{Result of Comparison}
	\label{fig:svd}
\end{figure}

We conduct experiments on real-world datasets described in Table~\ref{tb:data} . 
The sizes of these datasets range from thousands to millions.  
The datasets include both dense and sparse data matrices.
In our experiments, we set the target rank $k = 10$ and set $c$ and $r$ are several times of $k$ in both Fast SP-SVD  and Practical SP-SVD (Algorithm~\ref{alg:sp_svd_hk} in Appendix).
In their implementation, to have a fair comparison, we both use Gaussian projection for dense input matrices and count sketch for sparse input matrices.
Furthermore, in the Fast SP-SVD, we set $s_c = 3c \cdot a^{1/2}$ and $s_r = 3r \cdot a^{1/2}$.
We report the error ratio against $(c+r)/k$ in Figure~\ref{fig:svd}.

From Figure~\ref{fig:svd}, we can see that our algorithm is substantially better than Practical SP-SVD in all datasets. 
Especially, when $(c+r)/k$ is small, that is the sketching sizes are small, our Fast SP-SVD achieves much lower error ratio.
This result is constant with our theoretical comparison in Section~\ref{subsec:comparison}. 
Furthermore, our algorithm has another very practical advantage over Practical SP-SVD which can not be revealed from Figures. 
Theorem~\ref{thm:err_bnd} shows that the sketch sizes $c$ and $r$ of Algorithm~\ref{alg:Pass_effi_SVD} should be the same. Hence, we only need to tune one parameter. In contrast, the algorithm of \citet{tropp2017practical} need to tune two parameters which are closely interrelated. In fact, it is very cost to find a good parameters for this algorithm.

\section{Conclusion}

In this paper, we propose fast generalized matrix regression algorithm utilizing sketching techniques. 
We provide a tight bound on the sketching sizes for the fast GMR which are of order $\epsilon^{-1/2}$ to achieve a $(1+\epsilon)$ error bound for a large group of GMR problems.
This kind of bound of the sketching size is unknown before and the empirical study also validates the tightness of this bound.
The fast GMR algorithm is then applied to the SPSD approximation and the single pass SVD algorithm.
Our fast GMR based SPSD approximation method can achieve much better performance than the existing methods both theoretically and empirically.
Similarly, our fast single pass SVD algorithm outperform the practical single pass SVD proposed recently.  

\bibliography{reference}
\bibliographystyle{apalike2}

\cleardoublepage
\appendix

\section{Proof of Theorem~\ref{thm:gmp_main}}

Before the proof of Theorem~\ref{thm:gmp_main}, we first give several important lemmas which will be used in our proof.
First, we have the following lemma.
\begin{lemma}\label{lem:sep}
	We are given $\A\in \RR^{m\times n}$, $\C\in\RR^{m\times c}$ and $\R\in \RR^{r\times n}$. $\X^\star$ is defined as 
	\begin{equation}
	\X^\star = \C^{\dagger}\A\R^{\dagger}. \label{eq:U*}
	\end{equation} $\TX$ is a $c\times r$ matrix. Then we have
	\begin{equation*}
	\|\A - \C\TX\R\|_F^2 = \|\A - \C\X^\star\R\|_F^2 + \|\C(\X^\star-\TX)\R\|_F^2. 
	\end{equation*}
\end{lemma}
\begin{proof}
	We have
	\begin{align}
	&\|\A-\C\TX\R\|_F^{2}  \notag\\
	=& \|\A-\C\X^\star\R+\C\X^\star\R-\C\TX\R\|_F^2 \notag\\
	=& \|\A-\C\X^\star\R\|_F^{2}+\|\C\X^\star\R-\C\TX\R\|_F^{2} + 2\tr[(\A-\C\X^\star\R)^T(\C\X^\star\R-\C\TX\R)]  \notag\\
	=& \|\A-\C\X^\star\R\|_F^{2} + \|\C(\X^\star - \TX)\R\|_F^{2}, \notag
	\end{align}
	where the last equality is because
	\begin{align*}
	&\tr[(\A-\C\X^\star\R)^T(\C\X^\star\R-\C\TX\R)] \\
	=& \tr[((\I_m-\C\C^{\dagger})\A+(\C\C^{\dagger})\A(\I_n-\R^\dagger\R))^T\C(\X^\star-\TX)\R]\\
	=& \tr[\A^{T}(\I_m-\C\C^{\dagger})\C(\X^\star-\TX)\R] + \tr[\R(\I_n-\R^{\dagger}\R)\A^T\C\C^{\dagger}\C(\X^\star-\TX)] \\
	=& 0.
	\end{align*}
\end{proof}
Note that Lemma~\ref{lem:sep} holds for any matrix $\TX$ of size $c\times r$. 
Since $\|\A - \C\X^\star\R\|_F^2$ is a constant scalar,  we only need to bound the value of $\|\C(\X^\star-\TX)\R\|_F^2 $. 
Moreover, we have the following identity.
\begin{lemma}\label{lem:err_F}
	Given $\A\in \RR^{m\times n}$, $\C\in\RR^{m\times c}$ and $\R\in \RR^{r\times n}$, we also assume that $\S_C\in \RR^{s_c\times m}$ and  $\S_R\in\RR^{\s_r\times n}$ are sketch matrices described in Table~\ref{tb:property_theo}. Let $\TX$ and $\X^\star$ be defined in Eqn.~\eqref{eq:U_hat} and Eqn.~\eqref{eq:U*} respectively, then we have
	\[
	\|\C\X^\star\R - \C\TX\R\|_F = \|[(\S_C\U_C)^T\S_C\U_C]^{-1} \U_{C}^T\S_{C}^T\S_{C}\A^{\vdash} \S_{R}^T\S_{R}\V_{R}[(\S_{R}\V_{R})^T(\S_{R}\V_{R})]^{-1}\|_F,
	\]
	where $\A^\vdash \triangleq \A-\C\X^\star\R$.
\end{lemma}
\begin{proof}
	First, we consider $\C\TX\R$. Let the condensed SVDs of $\C$ and $\R$ be respectively
	\[
	\C = \U_C\Si_{C}\V_C^T \qquad \text{and } \qquad \R = \U_R\Si_R\V_R^T.
	\]
	Then, we have 
	\begin{align}
	\C\TX\R = & \C(\S_{C}\C)^{\dagger}\S_{C}\A\S_{R}^T(\R\S_{R}^T)^{\dagger}\R \notag\\
	=& \U_C\Si_{C}\V_C^T(\S_C\U_C\Si_{C}\V_C^T)^\dagger\S_{C}\A\S_{R}^T(\U_R\Si_R\V_R^T\S_R^T)^\dagger\U_R\Si_R\V_R^T \notag\\
	=&\U_C\Si_{C}\V_C^T(\Si_{C}\V_C^T)^\dagger(\S_C\U_C)^\dagger\S_{C}\A\S_{R}^T(\V_R^T\S_R^T)^\dagger(\U_R\Si_R)^{\dagger}\U_R\Si_R\V_R^T \label{eq:pinv}\\
	=& \U_C(\S_C\U_C)^\dagger\S_{C}\A\S_{R}^T(\V_R^T\S_R^T)^\dagger\V_R^{T}, \notag
	\end{align}
	where \eqref{eq:pinv} is because $\Si_{C}\V_{C}^T$ is of full row rank and $\U_R\Si_R$ is of full column rank such that 
	\[
	\Si_{C}\V_{C}^T(\Si_{C}\V_{C}^T)^{\dagger} = \I
	\]
	and
	\[
	(\U_R\Si_R)^{\dagger} \U_R\Si_R= \I.
	\]
	
	Similarly, we have 
	\begin{align*}
	\C\X^\star\R = \C\C^{\dagger}\A \R^{\dagger}\R = \U_{C}\U_{C}^T\A\V_{R}\V_{R}^T.
	\end{align*}
	We define 
	\[
	\A^\vdash \triangleq \A-\C\X^\star\R.
	\]
	Then, we obtain
	\begin{align*}
	&\C\X^\star\R - \C\TX\R\\
	=& \U_C(\S_C\U_C)^\dagger\S_{C}\A\S_{R}^T(\V_R^T\S_R^T)^\dagger\V_R^{T} - \U_{C}\U_{C}^T\A\V_{R}\V_{R}^T\\
	=& \U_C(\S_C\U_C)^\dagger(\S_{C}\A\S_{R}^T - \S_{C}\U_{C}\U_{C}^T\A\V_{R}\V_{R}^T\S_{R}^T)(\V_R^T\S_R^T)^\dagger\V_R^{T}\\
	=& \U_C(\S_C\U_C)^\dagger\S_{C}\A^{\vdash}\S_{R}^T(\V_R^T\S_R^T)^\dagger\V_R^{T}.
	\end{align*}
	Since  $\S_C$ and  $\S_R$ are the sketching matrices in Table~\ref{tb:property_theo}, then $\S_C\U_C$ is still of column full rank. 
	Thus, we have the following identities
	\begin{align*}
	(\S_C\U_C)^\dagger = [(\S_C\U_C)^T\S_C\U_C]^{-1}(\S_C\U_C)^T
	\end{align*}
	and 
	\begin{align*}
	(\V_R^T\S_R^T)^\dagger = (\V_R^T\S_R^T)^T[(\V_R^T\S_R^T)(\V_R^T\S_R^T)^T]^{-1}.
	\end{align*}
	Therefore, we have
	\begin{align*}
	\|\C\X^\star\R - \C\TX\R\|_F = \|[(\S_C\U_C)^T\S_C\U_C]^{-1} \U_{C}^T\S_{C}^T\S_{C}\A^{\vdash} \S_{R}^T\S_{R}\V_{R}[(\S_{R}\V_{R})^T(\S_{R}\V_{R})]^{-1}\|_F.
	\end{align*}	
\end{proof}
With Lemma~\ref{lem:err_F} at hand, we will show that $\C\X^\star\R$ and $\C\TX\R$ are close to each other. 
We first list two assumptions to help the proof of Theorem~\ref{thm:gmp_main}. 
In fact, these assumptions hold for the sketching matrices listed in Table~\ref{tb:property_theo}. 

\begin{assumption}
	\label{ass:Sc}
	Let $\Sc$ be a sketching matrix for $\C$ and $\C = \U_C\bSigma_C\V_C^T$ be the SVD decomposition of $\C$. 
	Let $\B$ be any matrix independent of $\Sc$ and be of consistent sizes with $\U_C^T$. 
	We assume that $\Sc$ satisfies that
	\begin{align}
	&\Pr\left\{
	\norm{\U_C^T\Sc^T\Sc\U_C - \I}_2 \ge 0.5
	\right\}
	\leq 
	\delta_1
	\label{eq:delta_U1}
	\\&
	\Pr\left\{
	\norm{\U_C^T\Sc^T\Sc\B - \U_C^T\B} \geq \epsilon_1 \norm{\U_C}_F\norm{\B}_F
	\right\}
	\leq
	\delta_2
	\label{eq:delta_U2}
	\end{align}
	with $\epsilon_1 \in (0, 1/2)$ and $\delta_1,\delta_2\in(0, 1/2)$.
\end{assumption}

\begin{assumption}
	\label{ass:Sr}
	Let $\Sr$ be a sketching matrix for $\R^T$ and $\R = \U_R\bSigma_R\V_R^T$ be the SVD decomposition of $\R$. 
	Let $\B'$ be any matrix independent of $\Sr$ be of consistent sizes consistent with $\V_R^T$. 
	We assume that $\Sc$ satisfies that
	\begin{align}
	&\Pr\left\{
	\norm{\V_R^T\Sr^T\Sr\V_R - \I}_2 \ge 0.5
	\right\}
	\leq 
	\delta_1 
	\label{eq:delta_V1}
	\\&
	\Pr\left\{
	\norm{\V_R^T\Sr^T\Sr\B' - \V_R^T\B'} \geq \epsilon_2 \norm{\V_R}_F\norm{\B'}_F
	\right\}
	\leq
	\delta_2
	\label{eq:delta_V2}
	\end{align}
	with $\epsilon_2 \in (0, 1/2)$ and $\delta_1,\delta_2\in(0, 1/2)$.
\end{assumption}

\begin{lemma}
	\label{lem:err}
	Let $\A$, $\C$, $\R$ be the input matrices of Algorithm~\ref{alg:gmr}. 
	Matrices $\S_C\in \RR^{s_c\times m}$ and  $\S_R\in\RR^{\s_r\times n}$ satisfy the properties described in Assumption~\ref{ass:Sc} with $\epsilon_1 = \min\left(
	\frac{\epsilon^{1/4}}{c^{1/2}},
	\frac{\epsilon^{1/2}}{c^{1/2}} \cdot \rho
	\right)$ and Assumption~\ref{ass:Sr} with $\epsilon_2 = \min\left(
	\frac{\epsilon^{1/4}}{r^{1/2}},
	\frac{\epsilon^{1/2}}{r^{1/2}} \cdot \rho
	\right)$,
	where $\rho$ is defined in Eqn.~\eqref{eq:rho}.  
	Let $\TX$ and $\X^\star$ be defined in Eqn.~\eqref{eq:U_hat} and Eqn.~\eqref{eq:U*} respectively, then we have
	\begin{equation*}
	\|\C\X^\star\R - \C\TX\R\|_F \leq 8\sqrt{\epsilon}\|\A-\C\X^\star\R\|_F.
	\end{equation*}
\end{lemma}
\begin{proof}
	By Eqn.~\eqref{eq:delta_U1} and \eqref{eq:delta_V1}, we can obtain that 
	\begin{equation}
	\label{eq:sig_val}
	0.5\le \sigma_{\min}(\Sc\U_C)\le 1.5, 
	\quad\mbox{and}\quad 
	0.5\le \sigma_{\min}(\Sr\V_R)\le 1.5.
	\end{equation}
	Combining with the result of Lemma~\ref{lem:err_F}, we have
	\begin{align*}
	\|\C\X^\star\R - \C\TX\R\|_F
	\leq&
	\norm{\U_C^T\Sc^T\Sc \A^\vdash\Sr^T\Sr\V_R}_F \cdot \norm{[(\S_C\U_C)^T\S_C\U_C]^{-1}}_2\cdot \norm{[(\S_{R}\V_{R})^T(\S_{R}\V_{R})]^{-1}}_2 
	\\\overset{\eqref{eq:sig_val}}{\le}&
	4\norm{\U_C^T\Sc^T\Sc \A^\vdash\Sr^T\Sr\V_R}_F
	\\=&
	4\norm{\U_C^T\Sc^T\Sc (\A - \U_C\U_C^T\A\V_R\V_R^T)\Sr^T\Sr\V_R}_F,
	\end{align*}
	
	By Eqn.~\eqref{eq:delta_U2} with 
	$
	\epsilon_1 = \min\left(
	\frac{\epsilon^{1/4}}{c^{1/2}},
	\frac{\epsilon^{1/2}}{c^{1/2}} \cdot \rho
	\right)
	$
	we have
	\begin{align*}
	&\norm{
		\U_C^T\Sc^T\Sc (\A - \U_C\U_C^T\A\V_R\V_R^T)\Sr^T\Sr\V_R 
		-
		(\U_C^T\A - \U_C^T\A\V_R\V_R^T) \Sr^T\Sr\V_R 
	}_F 
	\\\le& 
	\min\left(
	\frac{\epsilon^{1/4}}{c^{1/2}},
	\frac{\epsilon^{1/2}}{c^{1/2}} \cdot \rho
	\right)
	\cdot
	\norm{\U_C}_F 
	\cdot 
	\norm{ (\A - \U_C\U_C^T\A\V_R\V_R^T)\Sr^T\Sr\V_R }_F
	\\=&
	\min\left(\epsilon^{1/4},
	\epsilon^{1/2}\rho
	\right)
	\norm{ (\A - \U_C\U_C^T\A\V_R\V_R^T)\Sr^T\Sr\V_R }_F.
	\end{align*}
	Thus, we can obtain that
	\begin{equation}
	\label{eq:t_1}
	\begin{aligned}
	&\norm{
		\U_C^T\Sc^T\Sc (\A - \U_C\U_C^T\A\V_R\V_R^T)\Sr^T\Sr\V_R 
	}_F 
	\\\le& 
	\norm{(\U_C^T\A - \U_C^T\A\V_R\V_R^T) \Sr^T\Sr\V_R }_F
	+
	\min\left(\epsilon^{1/4},
	\epsilon^{1/2}\rho
	\right)
	\norm{ (\A - \U_C\U_C^T\A\V_R\V_R^T)\Sr^T\Sr\V_R }_F.
	\end{aligned}
	\end{equation}
	Then, we will bound above terms individually.
	First, by Eqn.~\eqref{eq:delta_V2} with 
	$
	\epsilon_2 = \sqrt{\frac{\epsilon}{r}} \cdot \rho,
	$
	we have 
	\begin{align}
	&\norm{(\U_C^T\A - \U_C^T\A\V_R\V_R^T) \Sr^T\Sr\V_R} 
	\notag
	\\\le&
	\sqrt{\frac{\epsilon}{r}} 
	\cdot
	\rho \norm{\V_R}_F
	\cdot
	\norm{\U_C^T\A - \U_C^T\A\V_R\V_R^T}
	\notag
	\\=&
	\rho\sqrt{\epsilon}
	\cdot 
	\frac{\norm{\U_C^T\A - \U_C^T\A\V_R\V_R^T}_F}{\norm{\A - \U_C\U_C^T\A\V_R\V_R^T}_F}
	\cdot
	\norm{\A - \U_C\U_C^T\A\V_R\V_R^T}_F. \label{eq:t_2}
	\end{align}
	Moreover, by Eqn.~\eqref{eq:delta_V2} with 
	$
	\epsilon_2 = \frac{\epsilon^{1/4}}{r^{1/2}}, 
	$
	we have
	\begin{align*}
	&\norm{ (\A - \U_C\U_C^T\A\V_R\V_R^T)\Sr^T\Sr\V_R - (\A\V_R - \U_C\U_C^T\A\V_R)}_F
	\\\le&
	\frac{\epsilon^{1/4}}{r^{1/2}} \norm{\V_R}_F \norm{\A - \U_C\U_C^T\A\V_R\V_R^T}_F
	\\=&
	\epsilon^{1/4}\norm{\A - \U_C\U_C^T\A\V_R\V_R^T}_F
	\end{align*}
	Thus, we have
	\begin{align}
	&\norm{ (\A - \U_C\U_C^T\A\V_R\V_R^T)\Sr^T\Sr\V_R}_F
	\notag
	\\\le&
	\norm{\A\V_R - \U_C\U_C^T\A\V_R}_F 
	+
	\epsilon^{1/4}\norm{\A - \U_C\U_C^T\A\V_R\V_R^T}_F 
	\notag
	\\\le&
	\frac{\norm{\A\V_R - \U_C\U_C^T\A\V_R}_F }{\norm{\A - \U_C\U_C^T\A\V_R\V_R^T}_F } \cdot 
	\norm{\A - \U_C\U_C^T\A\V_R\V_R^T}_F 
	+
	\epsilon^{1/4}\norm{\A - \U_C\U_C^T\A\V_R\V_R^T}_F. \label{eq:t_3}
	\end{align}
	Combining Eqn.~\eqref{eq:t_1}, \eqref{eq:t_2}, and \eqref{eq:t_3}, we can obtain that
	\begin{align*}
	&\norm{
		\U_C^T\Sc^T\Sc (\A - \U_C\U_C^T\A\V_R\V_R^T)\Sr^T\Sr\V_R 
	}_F 
	\\\overset{\eqref{eq:t_2}}{\le}&
	\rho\sqrt{\epsilon}
	\frac{\norm{\U_C^T\A - \U_C^T\A\V_R\V_R^T}_F}{\norm{\A - \U_C\U_C^T\A\V_R\V_R^T}_F}
	\cdot
	\norm{\A - \U_C\U_C^T\A\V_R\V_R^T}_F
	\\&+
	\min\left(\epsilon^{1/4},
	\epsilon^{1/2}\rho
	\right)
	\norm{ (\A - \U_C\U_C^T\A\V_R\V_R^T)\Sr^T\Sr\V_R }_F
	\\\overset{\eqref{eq:t_3}}{\le}&
	\left( \sqrt{\epsilon}
	+
	\sqrt{\epsilon} \rho
	\cdot
	\frac{\norm{\A\V_R - \U_C\U_C^T\A\V_R}_F 
		+ 
		\norm{\U_C^T\A - \U_C^T\A\V_R\V_R^T}_F}{\norm{\A - \U_C\U_C^T\A\V_R\V_R^T}_F} 
	\right)
	\cdot
	\norm{\A - \U_C\U_C^T\A\V_R\V_R^T}_F
	\\=&
	2\sqrt{\epsilon} 
	\cdot
	\norm{\A - \U_C\U_C^T\A\V_R\V_R^T}_F,
	\end{align*}
	where the last equality is because of the definition of $\rho$.
	Thus, we have
	\begin{equation*}
	\|\C\X^\star\R - \C\TX\R\|_F
	\le
	8\sqrt{\epsilon} 
	\cdot
	\norm{\A - \U_C\U_C^T\A\V_R\V_R^T}_F.
	\end{equation*}
\end{proof}

Combining above lemmas, we can prove Theorem~\ref{thm:gmp_main} as follows.
\begin{proof}[Proof of Theorem~\ref{thm:gmp_main}]
	$\S_C$ and $\S_R$  are any sketch matrix in Table~\ref{tb:property_theo}. Then they satisfy Assumption~\ref{ass:Sc} with $\epsilon_1 = \min\left(
	\frac{\epsilon^{1/4}}{c^{1/2}},
	\frac{\epsilon^{1/2}}{c^{1/2}} \cdot \rho
	\right)$ and Assumption~\ref{ass:Sr} with $\epsilon_2 = \min\left(
	\frac{\epsilon^{1/4}}{r^{1/2}},
	\frac{\epsilon^{1/2}}{r^{1/2}} \cdot \rho
	\right)$, respectively. 
	Furthermore, probability parameters $\delta_1$ and $\delta_2$ of $\S_C$ and $\S_R$ are $0.01$. Then, Assumption~\ref{ass:Sc} and Assumption~\ref{ass:Sr} hold simultaneously with probability at least $(1-0.01)^4 = 0.96$.
	
	By Lemma~\ref{lem:err_F} and \ref{lem:err}, we can obtain that 
	\begin{align*}
	\|\A-\C\TX\R\|_F^{2}  = & \|\A-\C\X^\star\R\|_F^{2}+\|\C\X^\star\R-\C\TX\R\|_F^{2} \\
	\leq& \|\A-\C\X^\star\R\|_F^{2} + 64\epsilon\|\A-\C\X^\star\R\|_F^{2}\\
	=&(1+64\epsilon)\|\A-\C\X^\star\R\|_F^{2}.
	\end{align*}
	By rescaling $\epsilon$, we get the result.
	
	For time complexity, it takes $\TimeSketch$ to compute $\S_C\C$, $\R\S_R^T$ and $\S_C\A\S_R^T$. Computing the pseudo-inverse of $\S_C\C$ and $\R\S_R^T$ requires $\cO(s_cc^2)$ and $\cO(s_rr^2)$ respectively. 
	The multiplication of $(\S_C\C)^{\dagger}\S_C\A\S_R^T(\R\S_R^T)^{\dagger}$ costs $\cO(s_c s_r\min(c,r))$. Hence, the time complexity of constructing $\TX$ is
	\[
	\cO(s_rr^2 +s_c c^2 + s_cs_r\cdot \min(c,r)) + T_{\mathrm{sketch}}.
	\]
\end{proof}

\section{Proof of Theorem~\ref{thm:sym}}

\begin{proof}[Proof of Theorem~\ref{thm:sym}] 
	By Theorem~\ref{thm:gmp_main}, we have 
	\[
	\|\A - \C\TX\C^T\|_F \leq (1+\epsilon)\min_{\X}\|\A - \C\X\C^T\|_F.
	\]
	Since $\TX_{\sym}$ is the projection of $\TX$ onto $\HB^s$ (defined in Eqn.~\eqref{eq:pi_H}), by Proposition~\ref{fact:cvx_pro}, we have
	\begin{align*}
	\|\A - \C\TX_{\sym}\C^T\|_F \leq \|\A - \C\TX\C^T\|_F \leq (1+\epsilon)\min_{\X}\|\A - \C\X\C^T\|_F.
	\end{align*}
	
	Similarly, if $\A$ is  SPSD, since $\TX_+$ is the projection of $\TH$ onto $\HB_+^s$ (defined in Eqn.~\eqref{eq:pi_H_+}), $\TX_+$ satisfies
	\begin{align*}
	\|\A - \C\TX_+\C^T\|_F \leq \|\A - \C\TX\C^T\|_F \leq (1+\epsilon)\min_{\X}\|\A - \C\X\C^T\|_F.
	\end{align*} 
\end{proof}

\begin{algorithm}[tb]
	\caption{Practical Single Pass SVD by \citet{tropp2017practical,clarkson2013low}.}
	\label{alg:sp_svd_hk}
		\begin{algorithmic}[1]
			\STATE {\bf Input:} A real matrix $\A \in \RBmn$, target rank $k$ and sketching size $c$ and $r$;
			\STATE Construct sketching matrices $\tilde{\Ps}\in\RR^{r\times m}$,  $\tilde{\Ome}\in\RR^{n\times c}$.
			\STATE $\C = [\quad]$, $\R = [\quad]$.
			\WHILE {$\A$ is not completely read through}
			\STATE Read next $L$-columns of columns of $\A$ denoted by $\A_L$;
			\STATE Update $\R = [\R, \tilde{\Ps}\A_L]$ and $\C = \C + \A_L\tilde{\Ome}$; 
			\ENDWHILE
			\STATE Compute the orthonormal bases of $\C$ and $\V^T$ by $[\U_{C},\T_C] = \mathrm{qr}(\C,0)$ and $[\V_{\R},\T_R] = \mathrm{qr}(\R^T,0)$ respectively. 
			\STATE Compute $\N = (\tilde{\Ps}\U_C)^{\dagger}\R\V_R$. 
			\STATE Compute the SVD decomposition $[\U_{N}, \Si, \V_{N}] = \mathrm{svd}(\N)$. 
			\STATE {\bf Output:} $\U = \U_{C}\U_{N}$, $\Si$, and $\V = \V_{R}\V_{N}$.
		\end{algorithmic}
\end{algorithm}

\section{Proof of Theorem~\ref{thm:lw_psd}}

\begin{proof}[Proof of Theorem~\ref{thm:lw_psd}]
	The proof of $\|\K - \C\TX_+\C^T\|_F \leq (1+\epsilon) \min_{\X} \|\K - \C\X\C^T\|_F$ is almost the same to the one of Theorem~\ref{thm:sym}.
	Thus, we will omit the detailed proof.
	
	Next, we will analyze the query complexity of Algorithm~\ref{alg:psd}. First, it requires to query $nc$ entries of the kernel to construct $\C$.
	Next, to compute the core matrix, it needs $s\times s$ entries of the kernel, that is, $c^2\cdot\max\left\{\epsilon^{-1}, \; \epsilon^{-2}\rho^{-4}\right\}$.
	Therefore, the total query complexity is 
	\begin{equation*}
	N = nc + c^2\cdot\max\left\{\epsilon^{-1}, \; \epsilon^{-2}\rho^{-4}\right\}.
	\end{equation*}
\end{proof}

\section{Proof of Theorem~\ref{thm:err_bnd}}

Before the proof of Theorem~\ref{thm:err_bnd}, we introduce 
$\SF(\epsilon,k)$ projection which is a class of projection matrices which have a particular property defined as follows \citep{clarkson2017low}.
With the help of $\SF(\epsilon,k)$ projection, Theorem~\ref{thm:err_bnd} can be proved in a convenient way.

\begin{definition}[SF($\epsilon,k$) projection] 
	For given $\A\in\RBmn$, say that projection $\PP\in\RR^{m\times m}$ is left $\SF(\epsilon,k)$ for $A$ if
	\[
	\|(\I - \PP)\A\|_2^2\leq(\epsilon/k)\|\A-\A_k\|_F^2.
	\]
	Similarly, we say projection $\PP\in\RR^{n\times n}$ is right $\SF(\epsilon,k)$ for $A$ if
	\[
	\|\A(\I - \PP)\|_2^2\leq(\epsilon/k)\|\A-\A_k\|_F^2.
	\]
\end{definition}

By sketching matrices, we can construct SF($\epsilon,k$) projections very efficiently \citep{clarkson2017low}.

\begin{lemma}[\citep{clarkson2017low}] \label{lem:sf_sketch}
	Suppose $\S\in\RR^{s\times m}$ be an OSNAP matrix with $\cO\left(\frac{k}{\epsilon}\right)^{1+\gamma}$ and $\gamma$ being a positive constant or a Gaussian projection matrix with $s = \cO\left(\frac{k}{\epsilon}\right)$. Then $\PP_{\S\A}$ is right $\SF(\epsilon,k)$ for $\A\in\RBmn$ with probability at least $0.99$, where $\PP_{\S\A}$ is the orthonormal projection onto the row space of $\S\A$. Specifically, when $\S$ is OSNAP, then the number of non-zero entry per column $p$ is $\cO(1)$.
\end{lemma}

Let $\S_1$ and $\S_2$ be OSNAP and Gaussian projection matrix with proper dimension, respectively, then $\S_2\S_1$ still have the sketching property described in Lemma~\ref{lem:sf_sketch}.
\begin{lemma} \label{lem:comp_sf}
	$\S_1\in\RR^{s_1\times m}$ is an OSNAP matrix with $s_1 = \cO\left(\frac{k}{\epsilon}\right)^{1+\gamma}$, $\gamma$ being a positive constant, and the number of non-zero entry per column $p$ being $\cO(1)$. $\S_2\in\RR^{s_2\times s_1}$ is a Gaussian projection matrix with $s_1 = \cO\left(\frac{k}{\epsilon}\right)$. Then $\PP_{\S_2\S_1\A}$ is right $\SF(\epsilon,k)$ for $\A\in\RBmn$ with probability at least $0.98$, where $\PP_{\S_2\S_1\A}$ is the orthonormal projection onto the row space of $\S_2\S_1\A$. 
\end{lemma}

Besides OSNAP matrices and Gaussian projection, other sketching matrix such and sampling matrix can be used to construct $\SF(\epsilon,k)$ projection \citep{clarkson2017low}.

Now, we give the following lemma which is important to the proof of Theorem~\ref{thm:err_bnd}.
\begin{lemma}\label{lem:sf_prop}
	Let $\PP_1$ and $\PP_2$ be left and right $\SF(\epsilon,k)$ for $\A$. Then
	\[
	\|\A - \PP_1\A_k\PP_2\|_F^2\leq(1+4\epsilon)\|\A - \A_k\|_F^2.
	\]
\end{lemma}
\begin{proof}
	We have
	\begin{align*}
	\|\A - \PP_1\A_k\PP_2\|_F^2 =& \|\A - \A_k\|_F^2 + \|\A_k - \PP_1\A_k\PP_2\|_F^2+2\tr((\A-\A_k)^T(\A_k - \PP_1\A_k\PP_2))\\
	=&\|\A - \A_k\|_F^2 + \|\A_k - \PP_1\A_k\PP_2\|_F^2+ 2\tr((\A-\A_k)^T(\I-\PP_1)\A_k\PP_2),
	\end{align*}
	where the second equality is because $(\A-\A_k)^T\A_k = 0$.
	
	For $\|\A_k - \PP_1\A_k\PP_2\|_F^2$, by the fact $(\I - \PP_1)\PP_1 = 0$ and $\|\A\B\|_F \leq\|\A\|_2\|\B\|_F$, we have
	\begin{align*}
	\|\A_k - \PP_1\A_k\PP_2\|_F^2 =& \|\A_k - \PP_1\A_k + \PP_1\A_k - \PP_1\A_k\PP_2\|_F^2\\
	=&\|\A_k - \PP_1\A_k\|_F^2+\|\PP_1\A_k - \PP_1\A_k\PP_2\|_F^2 \\
	\leq&\|\A_k - \PP_1\A_k\|_F^2 + \|\A_k - \A_k\PP_2\|_F^2. 
	\end{align*}
	By the fact that $\rk(\A_k) = k$ and $\|\A_k\x\|_2\leq\|\A\x\|_2$ for all $\x$,
	\begin{align*}
	\|\A_k - \PP_1\A_k\|_F^2 \leq& k\|(\I-\PP_1)\A_k\|_2^2\\
	\leq&k\|(\I-\PP_1)\A\|_2^2\\
	\leq&\epsilon\|\A - \A_k\|_F^2,
	\end{align*}
	where the last inequality is because  $\PP_1$ is left $\SF(\epsilon,k)$ for $\A$.
	
	Similarly, we have
	\[
	\|\A_k - \A_k\PP_2\|_F^2 \leq \epsilon\|\A - \A_k\|_F^2.
	\]
	We also have that
	\begin{align*}
	\tr((\A-\A_k)^T(\I-\PP_1)\A_k\PP_2) \leq& \sum_{i=1}^{k}\sigma_i((\I-\PP_1)(\A - \A_k))\sigma_i((\I-\PP_1)\A_k\PP_2)\\
	\leq&k\|(\I-\PP_1)(\A - \A_k)\|_2 \|(\I-\PP_1)\A_k\PP_2\|_2\\
	\leq&k\|(\I-\PP_1)\A\|_2^2\\
	\leq&\epsilon\|\A-\A_k\|_F^2
	\end{align*}
	
	Combining above results, we have
	\begin{align*}
	\|\A - \PP_1\A_k\PP_2\|_F^2 \leq \|\A - \A_k\|_F^2 + 4\epsilon\|\A-\A_k\|_F^2.
	\end{align*}
\end{proof}

\begin{corollary}\label{cor:sf}
	Let $\PP_C$ and $\PP_R$ be left and right $\SF(\epsilon,k)$ for $\A$. Then Eqn.~\eqref{eq:X} holds.
\end{corollary}
\begin{proof}
	By Lemma~\ref{lem:sf_prop}, we have 
	\[
	\|\A - \PP_C\A_k\PP_R\|_F^2\leq(1+\cO(\epsilon))\|\A - \A_k\|_F^2.
	\]
	Combining the fact that $\min_{\X}\|\A - \U_C\X\V_R^T\|_F \leq \|\A - \PP_C\A_k\PP_R\|_F$, we obtain the Eqn.~\eqref{eq:X}.
\end{proof}

\begin{proof}[Proof of Theorem~\ref{thm:err_bnd}]
	We have $\C = \A\tilde{\Ome}$ and $\R = \ti{\Ps}\A$, where  $\tilde{\Ome} = \Ome^T\G_C^T$ and $\tilde{\Ps} = \G_R\Ps$. 
	$\Ome$ and $\Ps$ are OSNAP matrices and $\G_C$ and $\G_R$ are Gaussian projection matrices.
	By  Lemma~\ref{lem:comp_sf} and Corollary~\ref{cor:sf}, we have that $\U_C\U_C^T$ is left $\SF(\epsilon,k)$ for $\A$ with probability at least $0.98$ and $\V_{R}\V_{R}^T$ is right $\SF(\epsilon,k)$ for $\A$ with probability at least $0.98$.
	
	Hence, by Lemma~\ref{lem:sf_prop} it holds that
	\begin{align*}
	\|\A - \U_C\U_C^T\A\V_{R}\V_{R}^T\|_F^2 \leq \|\A - \U_C\U_C^T\A_k\V_{R}\V_{R}^T\|_F^2 \leq (1+4\epsilon)\|\A - \A_k\|_F^2.
	\end{align*}
	Combining Theorem~\ref{thm:gmp_main}, we have
	\begin{align*}
	\|\A - \U\Si\V^T\|_F^2 \leq(1+\epsilon)\|\A - \U_C\U_C^T\A\V_{R}\V_{R}^T\|_F^2\leq(1+5\epsilon)\|\A - \A_k\|_F^2.
	\end{align*} 
	By rescaling the $\epsilon$, we get the result.
	
	It holds with probability at least $0.98\times 0.98$ that $\U_C\U_C^T$ and $\V_{R}\V_{R}^T$ are $\SF(\epsilon,k)$ for $\A$. 
	Moreover, Theorem~\ref{thm:gmp_main} holds with probability at least $0.9$. Therefore, the success probability is at least $0.9$. 
	
	Now, we analyze the computational and space complexity of Algorithm~\ref{alg:Pass_effi_SVD}. 
	In the analysis, we will use the notation $\gamma$ which is a constant in $(0, 1)$ and related to the OSNAP sketching matrix depicted in Section~\ref{subsec:sketch}.
	The detailed analysis of computational cost is as follows:
	\begin{enumerate}
		\item The  costs of computing  $\C$ and $\R$ are  respectively  $\cO\left(\nnz(\A) + m\left(\frac{k}{\epsilon}\right)^{2+\gamma}\right)$ and $\cO\left(\nnz(\A) + n\left(\frac{k}{\epsilon}\right)^{2+\gamma}\right)$. Constructing $\M$ needs  $\cO\left(\nnz(\A)+\min\{m,n\}\cdot \max\left\{\frac{k}{\epsilon^{3/2}}, \frac{k}{\epsilon^2 \rho^2}\right\} +n\left(\frac{k}{\epsilon}\right)^{1+\gamma}\right)$ time.
		\item It requires $\cO\left(m\left(\frac{k}{\epsilon}\right)^2\right)$ and $\cO\left(n\left(\frac{k}{\epsilon}\right)^2\right)$ time to compute $\U_C$ and $\V_R$ by QR decomposition.
		\item The time complexity of obtaining $(\Sc\U_{C})^{\dagger}\M(\V_{R}^T\Sr^T)^{\dagger}$ is 
		$\cO\left((m+n) \frac{k}{\epsilon}
		+ \frac{k^2}{\epsilon^2}\cdot\max\left\{\frac{k}{\epsilon^{3/2}}, \frac{k}{\epsilon^2 \rho^2}\right\} +\left(\frac{k}{\epsilon}\right)^{3+2\gamma}\right)$. 
		The cost of SVD  of $(\Sc\U_{C})^{\dagger}\M(\V_{R}^T\Sr^T)^{\dagger}$ is $\cO\left(\left(\frac{k}{\epsilon}\right)^{3}\right)$.
		\item It costs $\cO\left(m\left(\frac{k}{\epsilon}\right)^2\right)$ and $\cO\left(n\left(\frac{k}{\epsilon}\right)^2\right)$ to obtain $\U$ and $\V$.
	\end{enumerate}
	Therefore, the total computational cost of Algorithm~\ref{alg:Pass_effi_SVD} is 
	\begin{equation}
	\cO\left(\nnz(\A) 
	+ 
	(m+n)\left(\frac{k}{\epsilon}\right)^{2+\gamma} 
	+
	\frac{k^2}{\epsilon^2}\cdot\max\left\{\frac{k}{\epsilon^{3/2}}, \frac{k}{\epsilon^2 \rho^2}\right\}
	+ 
	\left(\frac{k}{\epsilon}\right)^{3+2\gamma}\right). \label{eq:cc}
	\end{equation}

	Next, we  analyze the space complexity:
	\begin{enumerate}
		\item It needs $\cO(n)$ space to store $\Ps$ and $\Sc$, and  $\cO(m)$ to store $\Ome$ and $\Sr$. $\G_R$ and $\G_C$ need $\cO\left(\frac{k^{2+\gamma}}{\epsilon^{1+\gamma}}\right)$ space.
		\item It needs $\cO\left(m\left(\frac{k}{\epsilon}\right)\right)$ and $\cO\left(n\left(\frac{k}{\epsilon}\right)\right)$ space to store $\C$ and $\R$, respectively. The storage space of $\M$ requires $$\cO\left(\max\left\{\frac{k^2}{\epsilon^{3}}, \frac{k^2}{\epsilon^4 \rho^4}\right\} + \left(\frac{k}{\epsilon}\right)^{2(1+\gamma)}\right)$$.
	\end{enumerate}
	Therefore, the total space cost is 
	\[
	\cO\left((m+n)\frac{k}{\epsilon}+ \max\left\{\frac{k^2}{\epsilon^{3}}, \frac{k^2}{\epsilon^4 \rho^4}\right\} + \left(\frac{k}{\epsilon}\right)^{2+2\gamma}\right).
	\]
\end{proof}

\end{document}